\documentclass[10pt]{article} 
\usepackage[margin=1in]{geometry}

\usepackage{url} 
\usepackage[round]{natbib} 

\usepackage[utf8]{inputenc} 
\usepackage[T1]{fontenc}    
\usepackage[hidelinks]{hyperref}
\hypersetup{
    colorlinks,
    citecolor=blue,
    linkcolor=blue,
    filecolor=blue,      
    urlcolor=blue,
}
\usepackage{url}            
\usepackage{booktabs}       
\usepackage{amsfonts}       
\usepackage{nicefrac}       
\usepackage{microtype}      
\usepackage{xcolor}         

\usepackage{graphicx}
\usepackage{tikz}
\usepackage{stackengine}
\usepackage{enumitem}
\usepackage{makecell}
\usepackage{algorithm,algorithmic}
\usepackage{multirow}
\usepackage{subfiles}
\usepackage{bm}
\usepackage{amsmath}
\usepackage{amsthm}
\usepackage{array}

\newtheorem{assumption}{Assumption}
\newtheorem{lemma}{Lemma}

\newtheorem{theorem}{Theorem}

\newtheorem{definition}{Definition}

\newtheorem{remark}{Remark}

\makeatletter
\newsavebox\myboxA
\newsavebox\myboxB
\newlength\mylenA

\newcommand*\xbar[2][0.75]{%
    \sbox{\myboxA}{$\m@th#2$}%
    \setbox\myboxB\null
    \ht\myboxB=\ht\myboxA%
    \dp\myboxB=\dp\myboxA%
    \wd\myboxB=#1\wd\myboxA
    \sbox\myboxB{$\m@th\overline{\copy\myboxB}$}
    \setlength\mylenA{\the\wd\myboxA}
    \addtolength\mylenA{-\the\wd\myboxB}%
    \ifdim\wd\myboxB<\wd\myboxA%
       \rlap{\hskip 0.5\mylenA\usebox\myboxB}{\usebox\myboxA}%
    \else
        \hskip -0.5\mylenA\rlap{\usebox\myboxA}{\hskip 0.5\mylenA\usebox\myboxB}%
    \fi}
\makeatother

\renewcommand{\algorithmicrequire}{\textbf{Input:}}
\renewcommand{\algorithmicensure}{\textbf{Output:}}

\newcommand{\vect}[1]{\ensuremath{\mathbf{#1}}}

\newcommand{\argmin}{\mathop{\rm argmin}}
\newcommand{\argmax}{\mathop{\rm argmax}}

\newcommand{\td}{{\tt D}_{\w\w}}
\newcommand{\tdv}{{\tt D}_{\v\v}}

\newcommand{\FU}{F_{S^{\setminus}}}
\newcommand{\FS}{F_{S}}
\newcommand{\WF}{\widetilde F}
\newcommand{\WFS}{\widetilde F_{S}}
\newcommand{\WFU}{\widetilde F_{S^{\setminus}}}

\renewcommand{\v}{\bm{\nu}}

\newcommand{\w}{\bm{\omega}}

\newcommand{\wn}{\bm{\omega}_{S}^*}
\newcommand{\wnu}{\bm{\omega}_{S^{\setminus}}^*}
\newcommand{\hw}{\widehat{\w}}
\newcommand{\hwu}{\widehat{\w}_{S^{\setminus}}}
\newcommand{\hv}{\widehat{\v}}
\newcommand{\hvu}{\widehat{\v}_{S^{\setminus}}}
\newcommand{\tw}{\widetilde{\w}}
\newcommand{\tv}{\widetilde{\v}}
\newcommand{\bw}{\overline{\w}}
\newcommand{\bv}{\overline{\v}}

\newcommand{\vu}{\bm{\nu}^u}
\newcommand{\vuu}{\bm{\nu}^u_{S^{\setminus}}}
\newcommand{\vn}{\bm{\nu}_{S}^*}
\newcommand{\vnu}{\bm{\nu}_{S^{\setminus}}^*}

\newcommand{\wuu}{\bm{\omega}^u_{S^{\setminus}}}
\newcommand{\wu}{\bm{\omega}^u}

\newcommand{\x}{\vect{x}}

\newcommand{\VS}{{\tt V}_S}
\newcommand{\VSU}{{\tt V}_{S^{\setminus}}}

\newcommand{\partitle}[1]{\smallskip \noindent \textbf{#1.}}

\begin{document}

\title{Certified Minimax Unlearning with Generalization Rates and Deletion Capacity}
\date{}

\author{%
    Jiaqi Liu\textsuperscript{1},~~Jian Lou\textsuperscript{1,2,\textdagger},~~Zhan Qin\textsuperscript{1,\textdagger},~~Kui Ren\textsuperscript{1}\\
  \textsuperscript{1}Zhejiang University\\
  \textsuperscript{2}ZJU-Hangzhou Global Scientific and Technological Innovation Center\\
  \texttt{\{jiaqi.liu, jian.lou, qinzhan, kuiren\}@zju.edu.cn} \\
  \textsuperscript{\textdagger}Corresponding authors
}

\maketitle

\begin{abstract}
We study the problem of $(\epsilon,\delta)$-certified machine unlearning for minimax models. Most of the existing works focus on unlearning from standard statistical learning models that have a single variable and their unlearning steps hinge on the \emph{direct Hessian-based conventional Newton} update. We develop a new $(\epsilon,\delta)$-certified machine unlearning algorithm for minimax models. It proposes a minimax unlearning step consisting of a \emph{total Hessian-based complete Newton} update and the Gaussian mechanism borrowed from differential privacy. To obtain the unlearning certification, our method injects calibrated Gaussian noises by carefully analyzing the ``sensitivity'' of the minimax unlearning step (i.e., the closeness between the minimax unlearning variables and the retraining-from-scratch variables). We derive the generalization rates in terms of population strong and weak primal-dual risk for three different cases of loss functions, i.e., (strongly-)convex-(strongly-)concave losses. We also provide the deletion capacity to guarantee that a desired population risk can be maintained as long as the number of deleted samples does not exceed the derived amount. With training samples $n$ and model dimension $d$, it yields the order $\mathcal O(n/d^{1/4})$, which shows a strict gap over the baseline method of differentially private minimax learning that has $\mathcal O(n/d^{1/2})$. In addition, our rates of generalization and deletion capacity match the state-of-the-art rates derived previously for standard statistical learning models.
\end{abstract}
\section{Introduction}
Minimax models have been widely applied in a variety of machine learning applications, including generative adversarial networks \citep{goodfellow2014generative, arjovsky2017wasserstein}, adversarially robust learning \citep{madrytowards, sinhacertifying}, and reinforcement learning \citep{du2017stochastic, dai2018sbeed}. This is largely credited to the two-variable (i.e., primal and dual variables) model structure of minimax models, which is versatile enough to accommodate such diverse instantiations. As is common in machine learning practice, training a successful minimax model relies crucially on a potentially large corpus of training samples that are contributed by users. This raises privacy concerns for minimax models. Unlike standard statistical learning (STL) models, the privacy studies for minimax models are relatively newer. Most of the existing studies focus on privacy protection during the training phase under the differential privacy (DP) notion \citep{dwork2006calibrating} and federated minimax learning settings \citep{sharma2022federated}. Recent works in this direction have successfully achieved several optimal generalization performances measured in terms of the population primal-dual (PD) risk for DP minimax models specifically \citep{yang2022differentially, zhang2022bring, bassily2023differentially, boob2023optimal}.

\begin{table*}[ht]
    \centering
    \setlength{\belowcaptionskip}{0.15cm}
    \renewcommand{\arraystretch}{1.2}
    \caption{\it Summary of Results. Here (S)C means (strongly-)convex loss function, and (S)C-(S)C means (strongly-)convex-(strongly-)concave loss function. PD means Primal-Dual. $n$ is the number of training samples and $d$ is the model dimension.}
    \label{tab:1}
    \begin{tabular}{|c|c|c|c|c|}
    \hline
    Model & Unlearning Algorithm & Setting & Generalization Measure & Deletion Capacity \\ \hline
    \multirow{2}{*}{STL} & \begin{tabular}[c]{@{}c@{}}DP-based\\ \citep{bassily2019private}\end{tabular} & C & \multirow{2}{*}{\begin{tabular}[c]{@{}c@{}}Population Excess Risk\\ \citep{sekhari2021remember}\end{tabular}} & $\mathcal O(n/d^{1/2})$ \\ \cline{2-3} \cline{5-5} 
    & \citep{sekhari2021remember} & (S)C &   & $\mathcal O(n/d^{1/4})$  \\ \hline
    \multirow{5}{*}{\begin{tabular}[c]{@{}c@{}}Minimax\\Learning\end{tabular}} & \begin{tabular}[c]{@{}c@{}}DP-based\\ \citep{zhang2022bring}\end{tabular} & SC-SC & \multirow{3}{*}{\begin{tabular}[c]{@{}c@{}}Population Strong\\PD Risk\end{tabular}} & \multirow{3}{*}{$\mathcal O(n/d^{1/2})$} \\ \cline{2-3}
    & \begin{tabular}[c]{@{}c@{}}DP-based\\ \citep{bassily2023differentially}\end{tabular} & C-C  &   &  \\ \cline{2-5} 
    & \textbf{Our Work} & (S)C-(S)C & \begin{tabular}[c]{@{}c@{}}Population Weak or\\Strong PD Risk\end{tabular}  & $\mathcal O(n/d^{1/4})$ \\ \hline
    \end{tabular}
\end{table*}

Machine unlearning is an emerging privacy-respecting problem concerning already-trained models (i.e., during the post-training phase) \citep{cao2015towards, guo2019certified, sekhari2021remember, graves2021amnesiac, bourtoule2021machine, li2021online, shibata2021learning, wu2022puma, cheng2023gnndelete, chen2023boundary, tarun2023fast,wu2023deltaboost,ghazi2023ticketed,wang2023bfu}. That is, it removes certain training samples from the trained model upon their users' data deletion requests. It is driven by the right to be forgotten, which is mandated by a growing number of user data protection legislations enacted in recent years. Prominent examples include the European Union’s General Data Protection Regulation (GDPR) \citep{mantelero2013eu}, the California Consumer Privacy Act (CCPA), and Canada’s proposed Consumer Privacy Protection Act (CPPA). Machine unlearning comes with several desiderata. Besides sufficiently removing the influence of the data being deleted, it should be efficient and avoid the prohibitive computational cost of the baseline method to fully retrain the model on the remaining dataset from scratch. To guarantee the sufficiency of data removal, there are exact machine unlearning methods \citep{cao2015towards, ginart2019making, brophy2021machine, bourtoule2021machine, ullah2021machine, schelter2021hedgecut, chen2022graph, chen2022recommendation, yan2022arcane,DBLP:journals/corr/abs-2212-10717,xia2023equitable} and approximate machine unlearning methods \citep{golatkar2020eternal, wu2020deltagrad, golatkar2020forgetting, nguyen2020variational, neel2021descent, peste2021ssse, golatkar2021mixed, warnecke2021machine, izzo2021approximate, mahadevan2021certifiable, mehta2022deep, zhang2022prompt,291009,chien2022efficient,lin2023erm} (some can offer the rigorous $(\epsilon,\delta)$-certification \citep{guo2019certified, sekhari2021remember, suriyakumar2022algorithms, chien2023efficient} inspired by differential privacy). In addition, recent studies also point out the importance of understanding the relationship between the generalization performance and the amount of deleted samples \citep{sekhari2021remember, suriyakumar2022algorithms}. In particular, they introduce the definition of deletion capacity to formally quantify the number of samples that can be deleted for the after-unlearning model to maintain a designated population risk. However, most existing works so far have focused on machine unlearning for standard statistical learning models with one variable, which leaves it unknown how to design a machine unlearning method to meet all the desiderata above. 

Machine unlearning for minimax models becomes a pressing problem because the trained minimax models also have a heavy reliance on the training data, while the users contributing data are granted the right to be forgotten. In this paper, we study the machine unlearning problem for minimax models under the $(\epsilon,\delta)$-certified machine unlearning framework. We collect in Table \ref{tab:1} the results in this paper and comparisons with baseline methods that are adapted from previous papers to $(\epsilon,\delta)$-certified machine unlearning.

Our main contributions can be summarized as follows.
\begin{itemize}[leftmargin=*]
    \item \emph{Certified minimax unlearning algorithm:} We develop $(\epsilon,\delta)$-certified minimax unlearning algorithm under the setting of the strongly-convex-strongly-concave loss function. To sufficiently remove the data influence, the algorithm introduces the total Hessian consisting of both direct Hessian and indirect Hessian, where the latter is crucial to account for the inter-dependence between the primal and dual variables in minimax models. It leads to the complete Newton-based minimax unlearning update. Subsequently, we introduce the Gaussian mechanism from DP to achieve the $(\epsilon,\delta)$-minimax unlearning certification, which requires careful analysis for the closeness between the complete Newton updated variables and the retraining-from-scratch variables. 
    \item \emph{Generalization:} We provide generalization results for our certified minimax unlearning algorithm in terms of the population weak and strong primal-dual risk, which is a common generalization measure for minimax models. 
    \item \emph{Deletion capacity:} We establish the deletion capacity result, which guarantees that our unlearning algorithm can retain the generalization rates for up to $\mathcal O(n/d^{1/4})$ deleted samples. 
    It matches the state-of-the-art result under the standard statistical unlearning setting that can be regarded as a special case of our minimax setting. 
    \item \emph{Extension to more general losses:} We extend the certified minimax unlearning to more general loss functions, including convex-concave, strongly-convex-concave, and convex-strongly-concave losses, and provide the corresponding $(\epsilon,\delta)$-certification, population primal-dual risk, and deletion capacity results. 
    \item \emph{Extension with better efficiency:} We develop a more computationally efficient extension, which can also support successive and online deletion requests. It saves the re-computation of the total Hessian matrix during the unlearning phase, where the minimax unlearning update can be regarded as a total Hessian-based infinitesimal jackknife. It also comes with has slightly smaller population primal-dual risk though the overall rates of the risk and deletion capacity remain the same.
\end{itemize}
\section{Related work}
Machine unlearning receives increasing research attention in recent years, mainly due to the growing concerns about the privacy of user data that are utilized for machine learning model training. Since the earliest work by \cite{cao2015towards}, a variety of machine unlearning methods have been proposed, which can be roughly divided into two categories: exact unlearning and approximate unlearning.

\partitle{Exact machine unlearning} Methods for exact machine unlearning aim to produce models that perform identically to the models retrained from scratch.
Some exact unlearning methods are designed for specific machine learning models like k-means clustering \citep{ginart2019making} and random forests \citep{brophy2021machine}.
SISA \citep{bourtoule2021machine} proposes a general exact unlearning framework based on sharding and slicing the training data into multiple non-overlapping shards and training independently on each shard. During unlearning, SISA retrains only on the shards containing the data to be removed. 
GraphEraser \citep{chen2022graph} and RecEraser \citep{chen2022recommendation} further extend SISA to unlearning for graph neural networks and recommendation systems, respectively. 

\partitle{Approximate machine unlearning} 
Approximate machine unlearning methods propose to make a tradeoff between the exactness in data removal and computational/memory efficiency. Prior works propose diverse ways to update the model parameter and offer different types of unlearning certification. When it comes to the unlearning update, many existing works consider the Newton update-related unlearning step where the Hessian matrix of the loss function plays a key role \citep{guo2019certified,golatkar2020eternal,peste2021ssse,sekhari2021remember,golatkar2021mixed,mahadevan2021certifiable,suriyakumar2022algorithms,mehta2022deep,chien2023efficient}. This unlearning update is motivated by influence functions \citep{koh2017understanding}. In order to alleviate the computation of the Hessian,  \cite{golatkar2020eternal} and \cite{peste2021ssse} utilize Fisher Information Matrix to approximate the Hessian, mitigating its expensive computation and inversion. 
\cite{mehta2022deep} provide a variant of conditional independence coefficient to select sufficient sets for unlearning, avoiding the need to invert the entire Hessian matrix.
ML-forgetting \citep{golatkar2021mixed} trains a linear weights set on the core dataset which would not change by standard training and a linear weights set on the user dataset containing data to be forgotten. They use an optimization problem to approximate the forgetting Newton update. \cite{suriyakumar2022algorithms} leverage the proximal infinitesimal jackknife as the unlearning step in order to be applied to nonsmooth loss functions. In addition, they can achieve better computational efficiency and are capable of dealing with online delete requests. There are also many other designs achieving different degrees of speedup \citep{wu2020deltagrad, nguyen2020variational,  neel2021descent,zhang2022prompt}.

Apart from the various designs for the unlearning update, there are also different definitions of certified machine unlearning. 
Early works like \cite{guo2019certified} introduce a certified data-removal mechanism that adds random perturbations to the loss function at training time. \cite{golatkar2020eternal} introduce an information-theoretic-based certified unlearning notion and also add random noise to ensure the certification, which is specific to the Fisher Information Matrix and not general enough.
More recently, \cite{sekhari2021remember} propose the $(\epsilon,\delta)$-certified machine unlearning definition that does not require introducing additional randomization during training.    
More essential, \cite{sekhari2021remember} points out the importance of providing the generalization performance after machine unlearning. \cite{sekhari2021remember,suriyakumar2022algorithms} establish the generalization result in terms of the population risk and derive the deletion capacity guarantee.

However, most existing works only consider machine unlearning for STL models that minimize a single variable. None of the prior works provide ertified machine unlearning pertaining to minimax models, for which the generalization and deletion capacity guarantees are still unknown.
\section{Preliminaries and Baseline Solution}
\subsection{Minimax Learning}
The goal of minimax learning is to optimize the population loss $F(\w,\v)$, given by
\begin{equation}
    \label{eq:population}
    \min_{\w\in\mathcal W}\max_{\v\in\mathcal V} F(\w,\v):=\mathbb E_{z\sim\mathcal D}[f(\w,\v;z)],
\end{equation}
where $f:\mathcal{W}\times \mathcal{V}\times \mathcal{Z} \to \mathbb{R}$ is the loss function, $z\in\mathcal{Z}$ is a data instance from the distribution $\mathcal D$,  $\mathcal{W}\subseteq\mathbb{R}^{d_1}$ and $\mathcal{V}\subseteq\mathbb{R}^{d_2}$ are closed convex domains with regard to primal and dual variables, respectively. 
Since the data distribution $\mathcal D$ is unknown in practice, minimax learning turns to optimize the empirical loss $\FS(\w,\v)$, given by,
\begin{equation}
\label{eq.empirical}
    \min_{\w\in\mathcal W}\max_{\v\in\mathcal V} F_S(\w,\v):=\frac{1}{n}\sum^n_{i=1}f(\w,\v;z_i),
\end{equation}
where $S=\{z_1,\cdots,z_n\}$ is the training dataset with $z_i \sim \mathcal{D}$. 

We will consider $L$-Lipschitz, $\ell$-smooth and $\mu_{\w}$-strongly-convex-$\mu_{\v}$-strongly-concave loss functions, which are described in Assumption \ref{ass:1}\&\ref{ass:2} blow and more details can be found in Appendix \ref{appendix:A}.
\begin{assumption}
\label{ass:1}
    For any $z\in \mathcal Z$, the function $f(\w,\v;z)$ is $L$-Lipschitz and with $\ell$-Lipschitz gradients and $\rho$-Lipschitz Hessians on the closed convex domain $\mathcal W \times \mathcal V$. Moreover, $f(\w,\v;z)$ is convex on $\mathcal W$ for any $\v \in \mathcal V$ and concave on $\mathcal V$ for any $\w \in \mathcal W$.
\end{assumption}
\begin{assumption}
\label{ass:2}
    For any $z\in \mathcal Z$, the function $f(\w,\v;z)$ satisfies Assumtion \ref{ass:1} and $f(\w,\v;z)$ is $\mu_{\w}$-strongly convex on $\mathcal W$ for any $\v \in \mathcal V$ and $\mu_{\v}$-strongly concave on $\mathcal V$ for any $\w \in \mathcal W$.
\end{assumption}

Denote a randomized minimax learning algorithm by $A:\mathcal Z^n\rightarrow \mathcal W\times\mathcal V$ and its trained variables by $A(S)=(A_{\w}(S),A_{\v}(S))\in\mathcal W\times\mathcal V$. The generalization performance is a top concern of the trained model variables $(A_{\w}(S),A_{\v}(S))$ \citep{thekumparampil2019efficient, zhang2020newton, lei2021stability, farnia2021train, zhang2021generalization, zhang2022uniform, ozdaglar2022what}, which can be measured by population weak primal-dual (PD) risk or population strong PD risk, as formalized below.

\begin{definition}[\textbf{Population Primal-Dual Risk}]
    \label{def:PDrisks}
    The population weak PD risk of $A(S)$, $\triangle^w(A_{\w}(S),A_{\v}(S))$ and the population strong PD risk of $A(S)$, $\triangle^s(A_{\w}(S),A_{\v}(S))$ are defined as
    \begin{equation}
        \left \{
            \begin{aligned}
                & \triangle^w(A_{\w}(S),A_{\v}(S))=\max_{\v\in\mathcal V}\mathbb E[F(A_{\w}(S),\v)]-\min_{\w\in\mathcal W}\mathbb E[F(\w,A_{\v}(S))], \\  
                & \triangle^s(A_{\w}(S),A_{\v}(S))=\mathbb E[\max_{\v\in\mathcal V}F(A_{\w}(S),\v)-\min_{\w\in\mathcal W}F(\w,A_{\v}(S))].
            \end{aligned}
        \right .
    \end{equation}
\end{definition}
\partitle{Notations}
We introduce the following notations that will be used in the sequel. 
For a twice differentiable function $f$ with the arguments $\w \in \mathcal W$ and $\v \in \mathcal V$, we use $\nabla_{\w}f$ and $\nabla_{\v}f$ to denote the direct gradient of $f$ w.r.t. $\w$ and $\v$, respectively
and denote its Jacobian matrix as $\nabla f = [\nabla_{\w}f;\nabla_{\v}f]$. 
We use $\partial_{\w\w}f$, $\partial_{\w\v}f$, $\partial_{\v\w}f$, $\partial_{\v\v}f$ to denote the second order partial derivatives w.r.t. $\w$ and $\v$, correspondingly
and denote its Hessian matrix as $\nabla^2 f = [\partial_{\w\w}f, \partial_{\w\v}f; \partial_{\v\w}f, \partial_{\v\v}f]$.
We define the total Hessian of the function $f$ w.r.t. $\w$ and $\v$: $\mathtt{D}_{\w\w}f:=\partial_{\w\w}f-\partial_{\w\v}f\cdot\partial_{\v\v}^{-1}f\cdot\partial_{\v\w}f$ and 
    $\mathtt{D}_{\v\v}f:=\partial_{\v\v}f-\partial_{\v\w}f\cdot\partial_{\w\w}^{-1}f\cdot\partial_{\w\v}f$, 
where $\partial_{\v\v}^{-1}f$ and $\partial_{\w\w}^{-1}f$ are the shorthand of $(\partial_{\v\v}f(\cdot))^{-1}$ and $(\partial_{\w\w}f(\cdot))^{-1}$, respectively, when $\partial_{\v\v}f$ and $\partial_{\w\w}f$ are invertible. 
We also use the shorthand notation $\nabla_{\w}f(\w_1,\v;z)=\left .\nabla_{\w}f(\w,\v;z)\right |_{\w=\w_1}$.

\subsection{$(\epsilon,\delta)$-Certified Machine Unlearning}
An unlearning algorithm $\bar A$ for minimax models receives the output of a minimax learning algorithm $A(S)$, the set of delete requests $U\subseteq S$ and some additional memory variables $T(S)\in\mathcal T$ as input and returns an updated model $(\wu,\vu)=(\bar A_{\w}(U,A(S),T(S)),\bar A_{\v}(U,A(S),T(S)))\in\mathcal W\times\mathcal V$, aiming to remove the influence of $U$. For the memory variables in $T(S)$, it will not contain the entire training set, but instead its size $|T(S)|$ is independent of the training data size $n$. The mapping of an unlearning algorithm can be formulated as $\bar A: \mathcal Z^m\times\mathcal W\times\mathcal V\times\mathcal T\rightarrow\mathcal W\times \mathcal V$. We now give the notion of $(\epsilon,\delta)$-certified unlearning introduced by  \cite{sekhari2021remember}, which is inspired by the definition of differential privacy \citep{dwork2006calibrating}.
\begin{definition}[\textbf{$(\epsilon,\delta)$-Certified Unlearning  \citep{sekhari2021remember}}]
    \label{def:unlearning}
    Let $\varTheta$ be the domain of $\mathcal{W}\times \mathcal{V}$. For all $S$ of size $n$, set of delete requests $U\subseteq S$ such that $|U|\leq m$, the pair of learning algorithm $A$ and unlearning algorithm $\bar A$ is $(\epsilon,\delta)$-certified unlearning, if $\forall O\subseteq \varTheta$ and $\epsilon,\delta>0$, the following two conditions are satisfied:
    \begin{gather}
        \Pr[\bar A(U,A(S),T(S))\in O]\leq e^{\epsilon}\cdot\Pr[\bar A(\emptyset,A(S\backslash U),T(S\backslash U))\in O]+\delta,\\
        \Pr[\bar A(\emptyset,A(S\backslash U),T(S\backslash U))\in O] \leq e^{\epsilon}\cdot\Pr[\bar A(U,A(S),T(S))\in O]+\delta,
    \end{gather}
    where $\emptyset$ denotes the empty set and $T(S)$ denotes the memory variables available to $\bar A$.
\end{definition}
The above definition ensures the indistinguishability between the output distribution of (i) the model trained on the set $S$ and then unlearned with delete requests $U$ and (ii) the model trained on the set $S\backslash U$ and then unlearned with an empty set. Specifically, the unlearning algorithm simply adds perturbations to the output of $A(S\backslash U)$ when the set of delete requests is empty.

\partitle{Deletion Capacity} Under the definition of certified unlearning, \cite{sekhari2021remember} introduce the definition of deletion capacity, which formalizes how many samples can be deleted while still maintaining good guarantees on test loss. Here, we utilize the population primal-dual risk defined in Definition \ref{def:PDrisks} instead of the excess population risk utilized for STL models.

\begin{definition}[\textbf{Deletion capacity, \citep{sekhari2021remember}}]
    \label{def:delcap}
    Let $\epsilon,\delta,\gamma>0$ and $S$ be a dataset of size $n$ drawn i.i.d from the data distribution $\mathcal D$. Let $F(\w,\v)$ be a minimax model and $U$ be the set of deletion requests. For a pair of minimax learning algorithm $A$ and minimax unlearning algorithm $\bar A$ that satisfies $(\epsilon,\delta)$-unlearning, the deletion capacity $m^{A,\bar A}_{\epsilon,\delta,\gamma}(d_1,d_2,n)$ is defined as the maximum number of samples $U$ that can be unlearned while still ensuring the population primal-dual (weak PD or strong PD) risk is at most $\gamma$. Let the expectation $\mathbb E[\cdot]$ takes over $S\sim\mathcal D^n$ and the outputs of the algorithms $A$ and $\bar A$. Let $d_1$ denotes the dimension of domain $\mathcal W$ and $d_2$ denotes the dimension of domain $\mathcal V$, specifically,
    \begin{equation}
        m^{A,\bar A}_{\epsilon,\delta,\gamma}(d_1,d_2,n):=\max\left\{m|\triangle\left(\bar A_{\w}(U,A(S),T(S)),\bar A_{\v}(U,A(S),T(S))\right)\leq\gamma\right\},
    \end{equation}
    where the ouputs $\bar A_{\w}(U,A(S),T(S))$ and $\bar A_{\v}(U,A(S),T(S))$ of the minimax unlearning algorithm $\bar A$ refer to parameter $\w$ and $\v$, respectively. $\triangle\left(\bar A_{\w}(U,A(S),T(S)),\bar A_{\v}(U,A(S),T(S))\right)$ could be the population weak PD risk or population strong PD risk of $\bar A(U,A(S),T(S))$.
\end{definition}
We set $\gamma = 0.01$ (or any other small arbitrary constant) throughout the paper.

\subsection{Baseline Solution: Certified Minimax Unlearning via Differential Privacy}
Since Definition \ref{def:unlearning} is motivated by differential privacy (DP), it is a natural way to use tools from DP for machine unlearning. For a differentially private learning algorithm $A$ with edit distance $m$ in neighboring datasets, the unlearning algorithm $\bar A$ simply returns its output $A(S)$ without any changes and is independent of the delete requests $U$ as well as the memory variables $T(S)$, i.e., $\bar A(U,A(S),T(S))=A(S)$.

A number of differentially private minimax learning algorithms can be applied, e.g.,  \cite{zhang2022bring, yang2022differentially, bassily2023differentially}. For instance, we can obtain the output $A(S)=(A_{\w}(S),A_{\v}(S))$ by calling Algorithm 3 in \cite{zhang2022bring}. Under Assumption \ref{ass:1}\&\ref{ass:2}, we then get the population strong PD risk based on \cite[Theorem 4.3]{zhang2022bring} and the group privacy property of DP \citep[Lemma 7.2.2]{vadhan2017complexity}, as follows,
\begin{equation}
\label{eq:dp-pdrisk}
    \triangle^s (A_{\w}(S),A_{\v}(S)) = \mathcal O \left( \frac{\kappa^2}{\mu n} + \frac{m^2 \kappa ^2 d \log(m e^\epsilon / \delta)}{\mu n^2 \epsilon^2} \right),
\end{equation}
where we let $\mu = \min\{\mu_{\w},\mu_{\v}\}$, $\kappa = \ell / \mu$, $d=\max\{d_1,d_2\}$, and $m$ be the edit distance between datasets (i.e., the original dataset and the remaining dataset after removing samples to be forgotten).

The algorithm $A$ satisfies $(\epsilon,\delta)$-DP for any set $U\subseteq S$ of size $m$, that is,
\begin{equation}
        \Pr[A(S)\in O]\leq e^{\epsilon}\Pr[A(S\backslash U)\in O]+\delta \quad \text{and} \quad
        \Pr[A(S\backslash U)\in O]\leq e^{\epsilon}\Pr[A(S)\in O]+\delta.\nonumber
\end{equation}
Since we have $A(S)=\bar A(U,A(S),T(S))$ and $A(S\backslash U)=\bar A(\emptyset,A(S\backslash U),T(S\backslash U))$, the above privacy guarantee can be converted to the minimax unlearning guarantee in Definition \ref{def:unlearning}, implying that the pair $(A,\bar A)$ is $(\epsilon,\delta)$-certified minimax unlearning. According to Definition \ref{def:delcap}, the population strong PD risk in eq.(\ref{eq:dp-pdrisk}) yields the following bound on deletion capacity.
\begin{theorem}[\textbf{Deletion capacity of unlearning via DP \citep{sekhari2021remember}}]
    Denote $d=\max\{d_1,d_2\}$. There exists a polynomial time learning algorithm $A$ and unlearning algorithm $A$ for minimax problem of the form $\bar A(U,A(S),T(S))=A(S)$ such that the deletion capacity is:
    \begin{equation}\label{eq:deletioncap_DP}
        m ^ { A,\bar A} _ {\epsilon,\delta,\gamma}(d_1,d_2,n) \geq \widetilde \Omega \left( \frac{n\epsilon}{\sqrt{d\log(e^{\epsilon}/\delta)}}\right),
    \end{equation}
    where the constant in $\widetilde \Omega$-notation depends on the properties of the loss function $f$ (e.g., strongly convexity and strongly concavity parameters, Lipchitz continuity and smoothness parameters).
\end{theorem}
However, this DP minimax learning baseline approach provides an inferior deletion capacity. In the following sections, we show that the $d^{1/2}$ in the denominator of eq.(\ref{eq:deletioncap_DP})  can be further reduced to $d^{1/4}$.
\section{Certified Minimax Unlearning}
\label{sec.scsc}
In this section, we focus on the setting of the strongly-convex-strongly-concave loss function. We first provide the intuition for the design of the minimax unlearning step in Sec.\ref{subsec.intuition}, then provide the formal algorithm in Sec.\ref{subsec.minimax.unlearning.alg} and a more efficient extension in Sec.\ref{subsec.efficient.extension} with analysis of minimax unlearning certification, generalization result, and deletion capacity in Sec.\ref{subsec.analysis.eff}. We will provide extensions to more general loss settings in Sec.\ref{sec.extentions}. The proofs for the theorems presented in this and the next sections can be found in Appendix \ref{appendix:B} and \ref{appendix:C}, respectively.
\subsection{Intuition for Minimax Unlearning Update}
\label{subsec.intuition}
To begin with, we provide an informal derivation for minimax unlearning update to illustrate its design intuition. Given the training set $S$ of size $n$ and the deletion subset $U\subseteq S$ of size $m$, the aim is to approximate the optimal solution $(\wnu,\vnu)$ of the loss $\FU(\w,\v)$ on the remaining dataset $S\setminus U$, given by,
    \begin{equation}
    \label{eq:remain_optimal}
        (\wnu,\vnu) := \arg\min_{\w\in\mathcal W}\max_{\v\in\mathcal V}\{\FU(\w,\v):=\frac{1}{n-m}\sum_{z_i\in S\setminus U}f(\w,\v;z_i)\}.
    \end{equation}
Meanwhile, we have the optimal solution $(\wn,\vn)$ to the original loss $\FS(\w,\v)$ after minimax learning. Taking unlearning $\w$ for instance, by using a first-order Taylor expansion for $\nabla_{\w}\FU(\wnu,\vnu) = 0$ around $(\wn,\vn)$, we have 
    \begin{equation}
        \nabla_{\w} \FU(\wn,\vn) + \partial_{\w\w} \FU(\wn,\vn) (\wnu-\wn) + \partial_{\w\v} \FU(\wn,\vn) (\vnu-\vn) \approx 0.
    \end{equation}
Since $\wn$ is a minimizer of $\FS(\w,\v)$, from the first-order optimality condition, we can get $\nabla_{\w} \FU(\wn,\vn) = - \frac{1}{n-m} \sum_{z_i \in U} \nabla_{\w} f(\wn,\vn;z_i)$. Now given an auxiliary function $\VSU(\w) = \argmax_{\v\in\mathcal V} \FU(\w,\v)$ (more best response auxiliary functions are introduced in Appendix \ref{appendix:A}, Definition \ref{def.aux.functions}), we have $\vnu = \VSU(\wnu)$. We further get
    \begin{equation}
        \begin{aligned}
            \vnu-\vn & = [\VSU(\wnu) - \VSU(\wn)] + [\VSU(\wn) - \vn] \\
            & \stackrel{(i)} {\approx} \VSU(\wnu) - \VSU(\wn) 
            \stackrel{(ii)} {\approx} \left( \frac{\mathrm d \VSU(\w^*_{S})}{\mathrm d \w} \Big{|}_{\w = \w^*_{S}}\right)(\wnu-\wn) \\
            & \stackrel{(iii)} {\approx} - \partial_{\v\v}^{-1} \FU(\wn,\vn) \cdot \partial_{\v\w} \FU(\wn,\vn) \cdot (\wnu-\wn),
        \end{aligned}        
    \end{equation}
where the approximate equation $(i)$ leaving out the term $[\VSU(\wn) - \vn]$ which is bounded in Appendix \ref{appendix:A}, Lemma \ref{lemma:vsuwn_vn}, and does not affect the overall unlearning guarantee. The approximate equation $(ii)$ is the linear approximation step and is the response Jacobian of the auxiliary function $\VSU(\w)$. And the approximate equation $(iii)$ is due to the implicit function theorem. This gives that
    \begin{equation}
             \partial_{\w\w} \FU(\wn,\vn) (\wnu-\wn) + \partial_{\w\v} \FU(\wn,\vn) (\vnu-\vn) 
            = \td \FU(\wn,\vn)(\wnu-\wn),
    \end{equation}
which implies the following approximation of $\wnu$:
\begin{equation}
\label{eq.complete.newton}
    \wnu \approx \wn+\frac{1}{n-m}[\td \FU(\wn,\vn)]^{-1}\sum_{z_i\in U}\nabla_{\w} f(\wn,\vn;z_i).
\end{equation}
The above informal derivation indicates that the minimax unlearning update relies on the total Hessian to sufficiently remove the data influence \cite{Liu_2023_ICCV,zhang2023closed}, rather than the conventional Hessian that appears in standard statistical unlearning \citep{guo2019certified, sekhari2021remember, suriyakumar2022algorithms, mehta2022deep}. The update in eq.(\ref{eq.complete.newton}) has a close relation to the complete Newton step in the second-order minimax optimization literature \citet{zhang2020newton}, which motivates the complete Newton-based minimax unlearning. However, due to the various approximations in the above informal derivation, we cannot have a certified minimax unlearning guarantee. Below, we will formally derive the upper bound for these approximations in the closeness upper bound analysis. Based on the closeness upper bound, we will introduce the Gaussian mechanism to yield distribution indistinguishably result in the sense of $(\epsilon, \delta)$-certified minimax unlearning.

\subsection{Proposed Certified Minimax Unlearning}
\label{subsec.minimax.unlearning.alg}
We first provide algorithms under the setting of the smooth and strongly-convex-strongly-concave (SC-SC) loss function as described in Assumptions \ref{ass:1}\&\ref{ass:2}.

\paragraph{Minimax Learning algorithm.}
We denote our learning algorithm by $A_{sc-sc}$ and the pseudocode is shown in Algorithm \ref{alg:learning}. Given a dataset $S=\{z_i\}^n_{i=1}$ of size $n$ drawn independently from some distribution $\mathcal D$, algorithm $A_{sc-sc}$ computes the optimal solution $(\wn,\vn)$ to the empirical risk $\FS(\w,\v)$. $A_{sc-sc}$ then outputs the point $(\wn,\vn)$ as well as the additional memory variables $T(S):=\{\mathtt D_{\w\w}\FS(\wn,\vn), \mathtt D_{\v\v}\FS(\wn,\vn)\}$, which computes and stores the total Hessian of $\FS(\w,\v)$ at $(\wn,\vn)$.

\partitle{Minimax Unlearning Algorithm}
We denote the proposed certified minimax unlearning algorithm by $\bar A_{sc-sc}$ and present its pseudocode in Algorithm \ref{alg:unlearning}. Algorithm $\bar A_{sc-sc}$ takes the following inputs: the set of delete requests $U=\{z_j\}^m_{j=1}$ of size $m$, the trained minimax model $(\wn,\vn)$, and the memory variables $T(S)$. To have the certified minimax unlearning for $\w$, eq.(\ref{eq:remianTH_w}) computes the total Hessian of $\FU(\wn,\vn)$ by $\frac{n}{n-m}\td \FS(\wn,\vn)- \frac{1}{n-m}\sum _{z_i\in U} \td f(\wn,\vn,z_i)$, where the former term can be retrieved from the memory set and the latter is computed on the samples to be deleted; eq.(\ref{eq:def_hw}) computes the intermediate $\widehat\w$ by the complete Newton step based on the total Hessian $\td \FU(\wn,\vn)$; Line 3 injects calibrated Gaussian noise $\bm{\xi}_1$ to ensure $(\epsilon,\delta)$-certified minimax unlearning. The certified minimax unlearning for $\v$ is symmetric. We provide detailed analysis for Algorithm \ref{alg:unlearning} including minimax unlearning certification, generalization results,
and deletion capacity in Appendix \ref{appendix:B.0}.

\begin{algorithm}[htbp]
    \caption{Mimimax Learning Algorithm $(A_{sc-sc})$}
    \label{alg:learning}
    {\small
    \begin{algorithmic}[1]
    \renewcommand{\algorithmicrequire}{\textbf{Input:}}
    \renewcommand{\algorithmicensure}{\textbf{Output:}}
    \REQUIRE Dataset $S$ : $\{z_i\}^n_{i=1}\sim\mathcal D^n$, loss function: $f(\w,\v;z)$.
    \STATE Compute 
    \begin{equation}
        (\wn,\vn) \leftarrow \arg\min_{\w}\max_{\v} F_S(\w,\v) :=\frac{1}{n}\sum^n_{i=1}f(\w,\v;z_i).
    \end{equation}
    \ENSURE $(\wn, \vn, \mathtt D_{\w\w}\FS(\wn,\vn), \mathtt D_{\v\v}\FS(\wn,\vn))$

     \end{algorithmic}  
    }
\end{algorithm}

\begin{algorithm}[htbp]
    \caption{Certified Minimax Unlearning for Strongly-Convex-Strongly-Concave Loss $(\bar A_{sc-sc})$}
    \label{alg:unlearning}
    {\small
    \begin{algorithmic}[1]
    \renewcommand{\algorithmicrequire}{\textbf{Input:}}
    \renewcommand{\algorithmicensure}{\textbf{Output:}}
    \REQUIRE Delete requests $U$ : $\{z_j\}^m_{j=1}\subseteq S$, output of $A_{sc-sc}(S)$: $(\wn,\vn)$, loss function: $f(\w,\v;z)$, memory variables $T(S)$: $\{\mathtt D_{\w\w}\FS(\wn,\vn), \mathtt D_{\v\v}\FS(\wn,\vn)\}$, noise parameters: $\sigma_1$, $\sigma_2$.
    \STATE Compute
        \begin{equation}\label{eq:remianTH_w}
            \td \FU(\wn,\vn)=\frac{1}{n-m}\left(n\mathtt D_{\w\w}\FS(\wn,\vn)-\sum_{z_i\in U}\mathtt D_{\w\w}f(\wn,\vn;z_i)\right),
        \end{equation}  
        \begin{equation}
             \tdv \FU(\wn,\vn)=\frac{1}{n-m}\left(n\mathtt D_{\v\v}\FS(\wn,\vn)-\sum_{z_i\in U}\mathtt D_{\v\v}f(\wn,\vn;z_i)\right).
        \end{equation}
    \STATE Define
        \begin{equation}
        \label{eq:def_hw}
            \widehat\w=\wn+\frac{1}{n-m}[\td \FU(\wn,\vn)]^{-1} \sum_{z_i\in U}\nabla_{\w} f(\wn,\vn;z_i),
        \end{equation}
        \begin{equation}
        \label{eq:def_hv}
            \widehat\v=\vn+\frac{1}{n-m}[\tdv \FU(\wn,\vn)]^{-1} \sum_{z_i\in U}\nabla_{\v} f(\wn,\vn;z_i).
        \end{equation}
    \STATE $\wu=\widehat\w+\bm{\xi}_1$, where $\bm{\xi}_1\sim \mathcal N(0,\sigma ^2 _1\mathbf{I}_{d_1})$ and $\vu=\widehat\v+\bm{\xi}_2$, where $\bm{\xi}_2\sim \mathcal N(0,\sigma ^2 _2\mathbf{I}_{d_2})$.
    \ENSURE $(\wu,\vu)$.
     \end{algorithmic}  
     }
\end{algorithm}

\subsection{Certified Minimax Unlearning without Total Hessian Re-computation}
\label{subsec.efficient.extension}
We extend Algorithm \ref{alg:unlearning} and propose Algorithm \ref{alg:unlearning.extension.online} to reduce the computational cost of Algorithm \ref{alg:unlearning}. The complete Newton steps in eq.(\ref{eq:def_hw_eff}) and eq.(\ref{eq:def_hv_eff}) utilize the total Hessian $\td \FS(\wn,\vn)$ and $\tdv \FS(\wn,\vn)$ that are directly retrieved from the memory, rather than the updated total Hessian $\td \FU(\wn,\vn)$ and $\tdv \FU(\wn,\vn)$ used in Algorithm \ref{alg:unlearning}. The form in eq.(\ref{eq:def_hw_eff}) and eq.(\ref{eq:def_hv_eff}) can also be regarded as the total Hessian extension of the infinitesimal jackknife. In this way, it gets rid of the computationally demanding part of re-evaluating the total Hessian for samples to be deleted, which significantly reduces the computational cost. It turns out to be the same computational complexity as the state-of-the-art certified unlearning method developed for STL models \citep{suriyakumar2022algorithms}. Moreover, Algorithm \ref{alg:unlearning.extension.online} can be more appealing for the successive data deletion setting. 
\begin{algorithm}[htbp]
    \caption{Efficient Certified Minimax Unlearning $(\bar A_{\tt efficient})$}
    \label{alg:unlearning.extension.online}
   {\small
    \begin{algorithmic}[1]
    \renewcommand{\algorithmicrequire}{\textbf{Input:}}
    \renewcommand{\algorithmicensure}{\textbf{Output:}}
    \REQUIRE Delete requests $U$ : $\{z_j\}^m_{j=1}\subseteq S$, output of $A_{sc-sc}(S)$: $(\wn,\vn)$, loss function: $f(\w,\v;z)$, memory variables $T(S)$: $\{\mathtt D_{\w\w}\FS(\wn,\vn), \mathtt D_{\v\v}\FS(\wn,\vn)\}$, noise parameters: $\sigma_1$, $\sigma_2$.
    \STATE Compute
        \begin{equation}
        \label{eq:def_hw_eff}
            \tw=\wn+\frac{1}{n}[\td \FS(\wn,\vn)]^{-1}\sum_{z_i\in U}\nabla_{\w} f(\wn,\vn;z_i),
        \end{equation}
        \begin{equation}
        \label{eq:def_hv_eff}
            \tv=\vn+\frac{1}{n}[\tdv \FS(\wn,\vn)]^{-1}\sum_{z_i\in U}\nabla_{\v} f(\wn,\vn;z_i).
        \end{equation}
    \STATE $\tw^u=\tw+\bm{\xi}_1$, where $\bm{\xi}_1\sim \mathcal N(0,\sigma^2_1\mathbf{I}_{d_1})$ and $\tv^u=\tv+\bm{\xi}_2$, where $\bm{\xi}_2\sim \mathcal N(0,\sigma^2_2\mathbf{I}_{d_2})$ .
    \ENSURE $(\tw^u,\tv^u)$.
     \end{algorithmic}  
     }
\end{algorithm}
\subsection{Analysis for Algorithm \ref{alg:unlearning.extension.online}} 
\label{subsec.analysis.eff}
\partitle{$(\epsilon,\delta)$-Certificated Unlearning Guarantee}
The intermediate variables $(\tw,\tv)$ are distinguishable in distribution from the retraining-from-scratch variables $(\wnu,\vnu)$ because they are deterministic and the Taylor expansion introduces a certain amount of approximation. The following lemma quantifies the closeness between $(\tw,\tv)$ and $(\wnu,\vnu)$, which can be regarded as the ``sensitivity'' when applying the Gaussian mechanism. 
\begin{lemma}[\textbf{Closeness Upper Bound}]
\label{lem:sensitivity.eff}
    Suppose the loss function $f$ satisfies Assumption \ref{ass:1} and \ref{ass:2}, $\|\td \FS(\wn,\vn)\| \geq \mu_{\w\w}$ and $\| \tdv \FS(\wn,\vn) \| \geq \mu_{\v\v}$. Let $\mu = \min\{\mu_{\w},\mu_{\v},\mu_{\w\w},\mu_{\v\v}\}$. 
    Then, we have the closeness bound between $(\tw,\tv)$ in Line 1 of Algorithm \ref{alg:unlearning.extension.online} and $(\wnu,\vnu)$ in eq.(\ref{eq:remain_optimal}):
        \begin{equation}
            \{\| \wnu-\tw \|,\| \vnu-\tv \|\} \leq \frac{(8\sqrt{2}L^2 \ell^3 \rho / \mu^6 + 2\sqrt{2}L \ell^2 / \mu^3) m^2}{n^2}.
        \end{equation}
\end{lemma}
Equipped with Lemma \ref{lem:sensitivity}, we have the following certified unlearning guarantee by adding Gaussian noise calibrated according to the above closeness result. Due to the minimax structure, our analysis is more involved than the STL case \citep{ sekhari2021remember, suriyakumar2022algorithms}.

\begin{theorem} [\textbf{$(\epsilon,\delta)$-Minimax Unlearning Certification}]
\label{theorem:certi.eff}
    Under the same settings of Lemma \ref{lem:sensitivity.eff}, our minimax learning algorithm $A_{sc-sc}$ and unlearning algorithm $\bar A_{\tt efficient}$ is $(\epsilon,\delta)$-certified minimax unlearning if we choose
    \begin{equation}\label{eq:noise.eff_}
        \sigma_1 \text{~and~} \sigma_2 = \frac{2(8 \sqrt{2}L^2 \ell^3 \rho / \mu^6 + 2\sqrt{2}L \ell^2 / \mu^3) m^2}{n^2\epsilon} \sqrt{2\log(2.5/\delta)}.
    \end{equation}
\end{theorem}

\partitle{Generalization Guarantee}
Theorem \ref{theorem:guarantee_SC.eff} below provides the generalization result in terms of the population PD risk for the minimax unlearning algorithm $\bar A_{\tt efficient}$. 

\begin{theorem}[\textbf{Population Primal-Dual Risk}]
\label{theorem:guarantee_SC.eff}
    Under the same settings of Lemma \ref{lem:sensitivity.eff} and denote $d=\max\{d_1,d_2\}$,
    the population weak and strong PD risk for the certified minimax unlearning variables $(\tw^u,\tv^u)$ returned by Algorithm \ref{alg:unlearning.extension.online} are
    {\small
    \begin{equation}
    \left \{
        \begin{aligned}
            & \triangle^w (\tw^u,\tv^u) = \mathcal O \left( (L^3 \ell^3 \rho / \mu^6 + L^2 \ell^2/\mu^3)\cdot \frac{m^2 \sqrt{d\log(1/\delta)}}{n^2 \epsilon} + \frac{mL^2}{\mu n}\right),\\
            & \triangle^s (\tw^u,\tv^u) = \mathcal O \left( (L^3 \ell^3 \rho / \mu^6 + L^2 \ell^2/\mu^3)\cdot \frac{m^2 \sqrt{d\log(1/\delta)}}{n^2 \epsilon} + \frac{mL^2}{\mu n} + \frac{L^2\ell}{\mu^2 n}\right).            
        \end{aligned}
    \right .
    \end{equation}  
    }
\end{theorem}
\partitle{Deletion Capacity}
The population weak and strong PD risk given in Theorem \ref{theorem:guarantee_SC.eff} for the output of unlearning algorithms provides the following bound on deletion capacity.
\begin{theorem}[\textbf{Deletion Capacity}]
\label{theorem:del_cap.eff}
    Under the same settings of Lemma \ref{lem:sensitivity.eff} and denote $d=\max\{d_1,d_2\}$, the deletion capacity of Algorithm \ref{alg:unlearning.extension.online} is
    \begin{equation}
        m ^ { A,\bar A} _ {\epsilon,\delta,\gamma}(d_1,d_2,n) \geq c \cdot \frac{n\sqrt{\epsilon}}{(d\log(1/\delta))^{1/4}},
    \end{equation}
    where the constant $c$ depends on $L, l, \rho,$ and $\mu$ of the loss function $f$.
\end{theorem}
\section{Certified Minimax Unlearning for Convex-Concave Loss Function}
\label{sec.extentions}
We further extend the certified minimax unlearning for the convex-concave loss function. In addition, Appendix \ref{appendix:C} will provide the extension to convex-strongly-concave and strongly-convex-concave loss functions. Give the convex-concave loss function $f(\w,\v;z)$, similar to the unlearning for STL models \citep{sekhari2021remember}, we define the regularized function as $\widetilde f(\w,\v;z) = f(\w,\v;z) + \frac{\lambda}{2}\|\w\|^2 - \frac{\lambda}{2}\|\v\|^2$. Suppose the function $f$ satisfies Assumption \ref{ass:1}, then the function $\widetilde f$ is $\lambda$-strongly convex in $\w$, $\lambda$-strongly concave in $\v$, $(2L+\lambda\|\w\|+\lambda\|\v\|)$-Lipschitz, $\sqrt{2}(2\ell+\lambda)$-gradient Lipschitz and $\rho$-Hessian Lipschitz. It suffices to apply the minimax learning and unlearning algorithms in Sec.\ref{sec.scsc} to the regularized loss function with a properly chosen $\lambda$. We denote the learning and unlearning algorithms for convex-concave losses as $A_{c-c}$ and $\bar A_{c-c}$. Their implementation details are given in Appendix \ref{appendix:C}.
We suppose the SC-SC regularization parameter $\lambda$ satisfies $\lambda<\ell$.
Theorem \ref{theorem:guarantee_C} below summarizes guarantees of $(\epsilon,\delta)$-certified unlearning and population primal-dual risk (weak and strong) for Algorithm $\bar A_{c-c}$. 
\begin{theorem}
\label{theorem:guarantee_C}
    Let Assumption \ref{ass:1} hold and $d=\max\{d_1,d_2\}$. Suppose the parameter spaces $\mathcal W$ and $\mathcal V$ are bounded so that $\max_{\w\in\mathcal W}\Vert\w\Vert\leq B_{\w}$ and $\max_{\v\in\mathcal V}\Vert\v\Vert\leq B_{\v}$. We have,
    \begin{enumerate}[label=$(\alph*)$,leftmargin=*]
    \item \textbf{$(\epsilon,\delta)$-Minimax Unlearning Certification:} Our minimax learning algorithm $A_{c-c}$ and unlearning algorithm $\bar A_{c-c}$ is $(\epsilon,\delta)$-certified minimax unlearning.
    \item \textbf{Population Weak PD Risk:} The population weak PD risk for $(\wu,\vu)$ by algorithm $\bar A_{c-c}$ is 
        \begin{equation}
            \triangle^w(\wu,\vu) \leq \mathcal O \bigg( (L^3 \ell^3 \rho / \lambda^6 + L^2 \ell^2/\lambda^3)\cdot \frac{m^2 \sqrt{d\log(1/\delta)}}{n^2 \epsilon} + \frac{mL^2}{\lambda n} + \lambda(B_{\w}^2 + B_{\v}^2) \bigg).
        \end{equation}
    
    In particular, by setting $\lambda$ below
    \begin{equation}
        \lambda = \max \bigg\{ \frac{L}{\sqrt{B_{\w}^2+B_{\v}^2}}\sqrt{\frac{m}{n}}, \left(\frac{L^2 \ell^2 m^2 \sqrt{d\log(1/\delta)}}{(B_{\w}^2 + B_{\v}^2)n^2 \epsilon}\right)^{1/4}, \left(\frac{L^3 \ell^3 \rho m^2 \sqrt{d\log(1/\delta)}}{(B_{\w}^2 + B_{\v}^2)n^2 \epsilon}\right)^{1/7} \bigg\},
    \end{equation}
    
    we have the following population weak PD risk,
    \begin{equation}
        \triangle^w(\wu,\vu) \leq \mathcal O\bigg( c_1 \sqrt{\frac{m}{n}} + c_2 \big(\frac{d\log(1/\delta)}{\epsilon^2}\big)^{1/8} \sqrt{\frac{m}{n}} + c_3 \big( \frac{\sqrt{d\log(1/\delta)}}{\epsilon}\big)^{1/7}(\frac{m}{n})^{2/7} \bigg),
    \end{equation}
    
    where $c_1, c_2, c_3$ are constants that depend only on $L, l, \rho, B_{\w}$ and $B_{\v}$.
    \item \textbf{Population Strong PD Risk:} The population strong PD risk for $(\wu,\vu)$ by algorithm $\bar A_{c-c}$ is
        \begin{equation}
            \triangle^s(\wu,\vu) \leq \mathcal O \bigg( (L^3 \ell^3 \rho / \lambda^6 + L^2 \ell^2/\lambda^3)\cdot \frac{m^2 \sqrt{d\log(1/\delta)}}{n^2 \epsilon} + \frac{mL^2}{\lambda n} + \frac{L^2 \ell}{\lambda^2 n} + \lambda(B_{\w}^2 + B_{\v}^2) \bigg).
        \end{equation}
    
    In particular, by setting $\lambda$ below
    \begin{equation}
    \begin{aligned}
        \lambda = \max \bigg\{ & \frac{L}{\sqrt{B_{\w}^2+B_{\v}^2}}\sqrt{\frac{m}{n}}, \left( \frac{L^2 \ell}{(B_{\w}^2+B_{\v}^2) n} \right)^{1/3}, \left(\frac{L^2 \ell^2 m^2 \sqrt{d\log(1/\delta)}}{(B_{\w}^2 + B_{\v}^2)n^2 \epsilon}\right)^{1/4}, \\
        & \left(\frac{L^3 \ell^3 \rho m^2 \sqrt{d\log(1/\delta)}}{(B_{\w}^2 + B_{\v}^2)n^2 \epsilon}\right)^{1/7} \bigg\},
    \end{aligned}
    \end{equation}
    
    we have the following population strong PD risk,
    \begin{equation}
        \triangle^s(\wu,\vu) \leq \mathcal O\bigg( c_1 \sqrt{\frac{m}{n}} + c_2\frac{1}{\sqrt[3]{n}} + c_3 \big(\frac{d\log(1/\delta)}{\epsilon^2}\big)^{1/8} \sqrt{\frac{m}{n}} + c_4 \big( \frac{\sqrt{d\log(1/\delta)}}{\epsilon}\big)^{1/7}(\frac{m}{n})^{2/7} \bigg),
    \end{equation}    
    where $c_1, c_2, c_3, c_4$ are constants that depend only on $L, l, \rho, B_{\w}$ and $B_{\v}$.
    \item \textbf{Deletion Capacity:} The deletion capacity of Algorithm $\bar A_{c-c}$ is
    \begin{equation}
        m ^ { A,\bar A} _ {\epsilon,\delta,\gamma}(d_1,d_2,n) \geq c \cdot \frac{n\sqrt{\epsilon}}{(d\log(1/\delta))^{1/4}},
    \end{equation}
    where the constant $c$ depends on the constants $L, l,\rho, B_{\w}$ and $B_{\v}$. 
    \end{enumerate}
\end{theorem}
\section{Conclusion}
In this paper, we have studied the certified machine unlearning for minimax models with a focus on the generalization rates and deletion capacity, while existing works in this area largely focus on standard statistical learning models. We have provided a new minimax unlearning algorithm composed of the total Hessian-based complete Newton update and the Gaussian mechanism-based perturbation, which comes with rigorous $(\epsilon,\delta)$-unlearning certification. We have established generalization results in terms of the population weak and strong primal-dual risk and the correspondingly defined deletion capacity results for the strongly-convex-strongly-concave loss functions, both of which match the state-of-the-art rates obtained for standard statistical learning models. We have also provided extensions to other loss types like the convex-concave loss function. In addition, we have provided a more computationally efficient extension by getting rid of the total Hessian re-computation during the minimax unlearning phase, which can be more appealing for the successive data deletion setting. Although our bound for deletion capacity is better than that of DP by an order of $d^{1/4}$ and matches the state-of-the-art result established for unlearning under the STL setting, it remains unclear whether this bound is tight or not. In future work, we plan to extend to more general settings like the nonconvex-nonconcave loss function setting.

\section*{Acknowledgements}
    We would like to thank Gautam Kamath for his valuable comments on the presentation of the results in the previous version of the paper. Additionally, we extend our thanks to the reviewers and area chair of NeurIPS 2023 for their constructive comments and feedback. This work was supported in part by the National Natural Science Foundation of China (62072395, 62206207, U20A20178), and the National Key Research and Development Program of China (2020AAA0107705, 2021YFB3100300).

\bibliographystyle{plainnat}
\bibliography{ref}

\begin{thebibliography}{69}
\providecommand{\natexlab}[1]{#1}
\providecommand{\url}[1]{\texttt{#1}}
\expandafter\ifx\csname urlstyle\endcsname\relax
  \providecommand{\doi}[1]{doi: #1}\else
  \providecommand{\doi}{doi: \begingroup \urlstyle{rm}\Url}\fi

\bibitem[Arjovsky et~al.(2017)Arjovsky, Chintala, and
  Bottou]{arjovsky2017wasserstein}
Martin Arjovsky, Soumith Chintala, and L{\'e}on Bottou.
\newblock Wasserstein generative adversarial networks.
\newblock In \emph{International conference on machine learning}, volume~70,
  pages 214--223. PMLR, 2017.

\bibitem[Bassily et~al.(2019)Bassily, Feldman, Talwar, and
  Thakurta]{bassily2019private}
Raef Bassily, Vitaly Feldman, Kunal Talwar, and Abhradeep~Guha Thakurta.
\newblock Private stochastic convex optimization with optimal rates.
\newblock In \emph{Advances in Neural Information Processing Systems},
  volume~32, pages 11279--11288, 2019.

\bibitem[Bassily et~al.(2023)Bassily, Guzm{\'{a}}n, and
  Menart]{bassily2023differentially}
Raef Bassily, Crist{\'{o}}bal Guzm{\'{a}}n, and Michael Menart.
\newblock Differentially private algorithms for the stochastic saddle point
  problem with optimal rates for the strong gap.
\newblock In \emph{Conference on Learning Theory}, volume 195, pages
  2482--2508. {PMLR}, 2023.

\bibitem[Boob and Guzm{\'a}n(2023)]{boob2023optimal}
Digvijay Boob and Crist{\'o}bal Guzm{\'a}n.
\newblock Optimal algorithms for differentially private stochastic monotone
  variational inequalities and saddle-point problems.
\newblock \emph{Mathematical Programming}, pages 1--43, 2023.

\bibitem[Bourtoule et~al.(2021)Bourtoule, Chandrasekaran, Choquette-Choo, Jia,
  Travers, Zhang, Lie, and Papernot]{bourtoule2021machine}
Lucas Bourtoule, Varun Chandrasekaran, Christopher~A Choquette-Choo, Hengrui
  Jia, Adelin Travers, Baiwu Zhang, David Lie, and Nicolas Papernot.
\newblock Machine unlearning.
\newblock In \emph{2021 IEEE Symposium on Security and Privacy}, pages
  141--159. IEEE, 2021.

\bibitem[Brophy and Lowd(2021)]{brophy2021machine}
Jonathan Brophy and Daniel Lowd.
\newblock Machine unlearning for random forests.
\newblock In \emph{International Conference on Machine Learning}, volume 139,
  pages 1092--1104. PMLR, 2021.

\bibitem[Cao and Yang(2015)]{cao2015towards}
Yinzhi Cao and Junfeng Yang.
\newblock Towards making systems forget with machine unlearning.
\newblock In \emph{2015 IEEE symposium on security and privacy}, pages
  463--480. IEEE, 2015.

\bibitem[Chen et~al.(2022{\natexlab{a}})Chen, Sun, Zhang, and
  Ding]{chen2022recommendation}
Chong Chen, Fei Sun, Min Zhang, and Bolin Ding.
\newblock Recommendation unlearning.
\newblock In \emph{Proceedings of the ACM Web Conference 2022}, pages
  2768--2777. {ACM}, 2022{\natexlab{a}}.

\bibitem[Chen et~al.(2022{\natexlab{b}})Chen, Zhang, Wang, Backes, Humbert, and
  Zhang]{chen2022graph}
Min Chen, Zhikun Zhang, Tianhao Wang, Michael Backes, Mathias Humbert, and Yang
  Zhang.
\newblock Graph unlearning.
\newblock In \emph{Proceedings of the 2022 ACM SIGSAC Conference on Computer
  and Communications Security}, pages 499--513. {ACM}, 2022{\natexlab{b}}.

\bibitem[Chen et~al.(2023)Chen, Gao, Liu, Peng, and Wang]{chen2023boundary}
Min Chen, Weizhuo Gao, Gaoyang Liu, Kai Peng, and Chen Wang.
\newblock Boundary unlearning: Rapid forgetting of deep networks via shifting
  the decision boundary.
\newblock In \emph{Proceedings of the {IEEE/CVF} Conference on Computer Vision
  and Pattern Recognition}, pages 7766--7775. {IEEE}, 2023.

\bibitem[Cheng et~al.(2023)Cheng, Dasoulas, He, Agarwal, and
  Zitnik]{cheng2023gnndelete}
Jiali Cheng, George Dasoulas, Huan He, Chirag Agarwal, and Marinka Zitnik.
\newblock Gnndelete: A general strategy for unlearning in graph neural
  networks.
\newblock In \emph{The Eleventh International Conference on Learning
  Representations}. OpenReview.net, 2023.

\bibitem[Chien et~al.(2023{\natexlab{a}})Chien, Pan, and
  Milenkovic]{chien2022efficient}
Eli Chien, Chao Pan, and Olgica Milenkovic.
\newblock Efficient model updates for approximate unlearning of
  graph-structured data.
\newblock In \emph{The Eleventh International Conference on Learning
  Representations}. OpenReview.net, 2023{\natexlab{a}}.

\bibitem[Chien et~al.(2023{\natexlab{b}})Chien, Pan, and
  Milenkovic]{chien2023efficient}
Eli Chien, Chao Pan, and Olgica Milenkovic.
\newblock Efficient model updates for approximate unlearning of
  graph-structured data.
\newblock In \emph{The Eleventh International Conference on Learning
  Representations}. OpenReview.net, 2023{\natexlab{b}}.

\bibitem[Dai et~al.(2018)Dai, Shaw, Li, Xiao, He, Liu, Chen, and
  Song]{dai2018sbeed}
Bo~Dai, Albert Shaw, Lihong Li, Lin Xiao, Niao He, Zhen Liu, Jianshu Chen, and
  Le~Song.
\newblock Sbeed: Convergent reinforcement learning with nonlinear function
  approximation.
\newblock In \emph{International Conference on Machine Learning}, volume~80,
  pages 1125--1134. PMLR, 2018.

\bibitem[Di et~al.(2023)Di, Douglas, Acharya, Kamath, and
  Sekhari]{DBLP:journals/corr/abs-2212-10717}
Jimmy~Z. Di, Jack Douglas, Jayadev Acharya, Gautam Kamath, and Ayush Sekhari.
\newblock Hidden poison: Machine unlearning enables camouflaged poisoning
  attacks.
\newblock In \emph{Advances in Neural Information Processing Systems},
  volume~37, 2023.

\bibitem[Du et~al.(2017)Du, Chen, Li, Xiao, and Zhou]{du2017stochastic}
Simon~S Du, Jianshu Chen, Lihong Li, Lin Xiao, and Dengyong Zhou.
\newblock Stochastic variance reduction methods for policy evaluation.
\newblock In \emph{International Conference on Machine Learning}, volume~70,
  pages 1049--1058. PMLR, 2017.

\bibitem[Dwork et~al.(2006)Dwork, McSherry, Nissim, and
  Smith]{dwork2006calibrating}
Cynthia Dwork, Frank McSherry, Kobbi Nissim, and Adam Smith.
\newblock Calibrating noise to sensitivity in private data analysis.
\newblock In \emph{Third Theory of Cryptography Conference}, volume 3876, pages
  265--284. Springer, 2006.

\bibitem[Dwork et~al.(2014)Dwork, Roth, et~al.]{dwork2014algorithmic}
Cynthia Dwork, Aaron Roth, et~al.
\newblock The algorithmic foundations of differential privacy.
\newblock \emph{Foundations and Trends{\textregistered} in Theoretical Computer
  Science}, 9:\penalty0 211--407, 2014.

\bibitem[Farnia and Ozdaglar(2021)]{farnia2021train}
Farzan Farnia and Asuman Ozdaglar.
\newblock Train simultaneously, generalize better: Stability of gradient-based
  minimax learners.
\newblock In \emph{International Conference on Machine Learning}, volume 139,
  pages 3174--3185. PMLR, 2021.

\bibitem[Ghazi et~al.(2023)Ghazi, Kamath, Kumar, Manurangsi, Sekhari, and
  Zhang]{ghazi2023ticketed}
Badih Ghazi, Pritish Kamath, Ravi Kumar, Pasin Manurangsi, Ayush Sekhari, and
  Chiyuan Zhang.
\newblock Ticketed learning--unlearning schemes.
\newblock In \emph{The Thirty Sixth Annual Conference on Learning Theory},
  pages 5110--5139. PMLR, 2023.

\bibitem[Ginart et~al.(2019)Ginart, Guan, Valiant, and Zou]{ginart2019making}
Antonio Ginart, Melody Guan, Gregory Valiant, and James~Y Zou.
\newblock Making ai forget you: Data deletion in machine learning.
\newblock In \emph{Advances in neural information processing systems},
  volume~32, pages 3513--3526, 2019.

\bibitem[Golatkar et~al.(2020{\natexlab{a}})Golatkar, Achille, and
  Soatto]{golatkar2020eternal}
Aditya Golatkar, Alessandro Achille, and Stefano Soatto.
\newblock Eternal sunshine of the spotless net: Selective forgetting in deep
  networks.
\newblock In \emph{Proceedings of the IEEE/CVF Conference on Computer Vision
  and Pattern Recognition}, pages 9304--9312, 2020{\natexlab{a}}.

\bibitem[Golatkar et~al.(2020{\natexlab{b}})Golatkar, Achille, and
  Soatto]{golatkar2020forgetting}
Aditya Golatkar, Alessandro Achille, and Stefano Soatto.
\newblock Forgetting outside the box: Scrubbing deep networks of information
  accessible from input-output observations.
\newblock In \emph{Computer Vision--ECCV 2020: 16th European Conference},
  volume 12374, pages 383--398. Springer, 2020{\natexlab{b}}.

\bibitem[Golatkar et~al.(2021)Golatkar, Achille, Ravichandran, Polito, and
  Soatto]{golatkar2021mixed}
Aditya Golatkar, Alessandro Achille, Avinash Ravichandran, Marzia Polito, and
  Stefano Soatto.
\newblock Mixed-privacy forgetting in deep networks.
\newblock In \emph{Proceedings of the IEEE/CVF Conference on Computer Vision
  and Pattern Recognition}, pages 792--801. {IEEE}, 2021.

\bibitem[Goodfellow et~al.(2014)Goodfellow, Pouget-Abadie, Mirza, Xu,
  Warde-Farley, Ozair, Courville, and Bengio]{goodfellow2014generative}
Ian Goodfellow, Jean Pouget-Abadie, Mehdi Mirza, Bing Xu, David Warde-Farley,
  Sherjil Ozair, Aaron Courville, and Yoshua Bengio.
\newblock Generative adversarial nets.
\newblock In \emph{Advances in Neural Information Processing Systems},
  volume~27, pages 2672--2680, 2014.

\bibitem[Graves et~al.(2021)Graves, Nagisetty, and Ganesh]{graves2021amnesiac}
Laura Graves, Vineel Nagisetty, and Vijay Ganesh.
\newblock Amnesiac machine learning.
\newblock In \emph{Proceedings of the AAAI Conference on Artificial
  Intelligence}, volume~35, pages 11516--11524. {AAAI} Press, 2021.

\bibitem[Guo et~al.(2020)Guo, Goldstein, Hannun, and Van
  Der~Maaten]{guo2019certified}
Chuan Guo, Tom Goldstein, Awni Hannun, and Laurens Van Der~Maaten.
\newblock Certified data removal from machine learning models.
\newblock In \emph{International Conference on Machine Learning}, volume 119,
  pages 3832--3842. {PMLR}, 2020.

\bibitem[Izzo et~al.(2021)Izzo, Smart, Chaudhuri, and Zou]{izzo2021approximate}
Zachary Izzo, Mary~Anne Smart, Kamalika Chaudhuri, and James Zou.
\newblock Approximate data deletion from machine learning models.
\newblock In \emph{International Conference on Artificial Intelligence and
  Statistics}, volume 130, pages 2008--2016. PMLR, 2021.

\bibitem[Koh and Liang(2017)]{koh2017understanding}
Pang~Wei Koh and Percy Liang.
\newblock Understanding black-box predictions via influence functions.
\newblock In \emph{International Conference on Machine Learning}, volume~70,
  pages 1885--1894. PMLR, 2017.

\bibitem[Lei et~al.(2021)Lei, Yang, Yang, and Ying]{lei2021stability}
Yunwen Lei, Zhenhuan Yang, Tianbao Yang, and Yiming Ying.
\newblock Stability and generalization of stochastic gradient methods for
  minimax problems.
\newblock In \emph{International Conference on Machine Learning}, volume 139,
  pages 6175--6186. PMLR, 2021.

\bibitem[Li et~al.(2021)Li, Wang, and Cheng]{li2021online}
Yuantong Li, Chi-Hua Wang, and Guang Cheng.
\newblock Online forgetting process for linear regression models.
\newblock In \emph{International Conference on Artificial Intelligence and
  Statistics}, pages 217--225. PMLR, 2021.

\bibitem[Lin et~al.(2023)Lin, Zhang, Chen, Chen, and Susilo]{lin2023erm}
Shen Lin, Xiaoyu Zhang, Chenyang Chen, Xiaofeng Chen, and Willy Susilo.
\newblock Erm-ktp: Knowledge-level machine unlearning via knowledge transfer.
\newblock In \emph{Proceedings of the IEEE/CVF Conference on Computer Vision
  and Pattern Recognition}, pages 20147--20155, 2023.

\bibitem[Lin et~al.(2020)Lin, Jin, and Jordan]{lin2020gradient}
Tianyi Lin, Chi Jin, and Michael Jordan.
\newblock On gradient descent ascent for nonconvex-concave minimax problems.
\newblock In \emph{International Conference on Machine Learning}, volume 119,
  pages 6083--6093. PMLR, 2020.

\bibitem[Liu et~al.(2023)Liu, Xue, Lou, Zhang, Xiong, and Qin]{Liu_2023_ICCV}
Junxu Liu, Mingsheng Xue, Jian Lou, Xiaoyu Zhang, Li~Xiong, and Zhan Qin.
\newblock Muter: Machine unlearning on adversarially trained models.
\newblock In \emph{Proceedings of the IEEE/CVF International Conference on
  Computer Vision}, pages 4892--4902, 2023.

\bibitem[Luo et~al.(2022)Luo, Li, and Chen]{luo2022finding}
Luo Luo, Yujun Li, and Cheng Chen.
\newblock Finding second-order stationary points in nonconvex-strongly-concave
  minimax optimization.
\newblock In \emph{Advances in Neural Information Processing Systems},
  volume~35, pages 36667--36679, 2022.

\bibitem[Madry et~al.(2018)Madry, Makelov, Schmidt, Tsipras, and
  Vladu]{madrytowards}
Aleksander Madry, Aleksandar Makelov, Ludwig Schmidt, Dimitris Tsipras, and
  Adrian Vladu.
\newblock Towards deep learning models resistant to adversarial attacks.
\newblock In \emph{The Sixth International Conference on Learning
  Representations}. OpenReview.net, 2018.

\bibitem[Mahadevan and Mathioudakis(2021)]{mahadevan2021certifiable}
Ananth Mahadevan and Michael Mathioudakis.
\newblock Certifiable machine unlearning for linear models.
\newblock \emph{arXiv preprint arXiv:2106.15093}, 2021.

\bibitem[Mantelero(2013)]{mantelero2013eu}
Alessandro Mantelero.
\newblock The eu proposal for a general data protection regulation and the
  roots of the ‘right to be forgotten’.
\newblock \emph{Computer Law \& Security Review}, 29\penalty0 (3):\penalty0
  229--235, 2013.

\bibitem[Mehta et~al.(2022)Mehta, Pal, Singh, and Ravi]{mehta2022deep}
Ronak Mehta, Sourav Pal, Vikas Singh, and Sathya~N Ravi.
\newblock Deep unlearning via randomized conditionally independent hessians.
\newblock In \emph{Proceedings of the IEEE/CVF Conference on Computer Vision
  and Pattern Recognition}, pages 10422--10431. {IEEE}, 2022.

\bibitem[Neel et~al.(2021)Neel, Roth, and Sharifi-Malvajerdi]{neel2021descent}
Seth Neel, Aaron Roth, and Saeed Sharifi-Malvajerdi.
\newblock Descent-to-delete: Gradient-based methods for machine unlearning.
\newblock In \emph{Algorithmic Learning Theory}, volume 132, pages 931--962.
  PMLR, 2021.

\bibitem[Nesterov and Polyak(2006)]{nesterov2006cubic}
Yurii Nesterov and Boris~T Polyak.
\newblock Cubic regularization of newton method and its global performance.
\newblock \emph{Mathematical Programming}, 108\penalty0 (1):\penalty0 177--205,
  2006.

\bibitem[Nguyen et~al.(2020)Nguyen, Low, and Jaillet]{nguyen2020variational}
Quoc~Phong Nguyen, Bryan Kian~Hsiang Low, and Patrick Jaillet.
\newblock Variational bayesian unlearning.
\newblock In \emph{Advances in Neural Information Processing Systems},
  volume~33, pages 16025--16036, 2020.

\bibitem[Ozdaglar et~al.(2022)Ozdaglar, Pattathil, Zhang, and
  Zhang]{ozdaglar2022what}
Asuman~E. Ozdaglar, Sarath Pattathil, Jiawei Zhang, and Kaiqing Zhang.
\newblock What is a good metric to study generalization of minimax learners?
\newblock In \emph{Advances in Neural Information Processing Systems},
  volume~35, pages 38190--38203, 2022.

\bibitem[Peste et~al.(2021)Peste, Alistarh, and Lampert]{peste2021ssse}
Alexandra Peste, Dan Alistarh, and Christoph~H Lampert.
\newblock {SSSE:} efficiently erasing samples from trained machine learning
  models.
\newblock \emph{arXiv preprint arXiv:2107.03860}, 2021.

\bibitem[Schelter et~al.(2021)Schelter, Grafberger, and
  Dunning]{schelter2021hedgecut}
Sebastian Schelter, Stefan Grafberger, and Ted Dunning.
\newblock Hedgecut: Maintaining randomised trees for low-latency machine
  unlearning.
\newblock In \emph{International Conference on Management of Data}, pages
  1545--1557. {ACM}, 2021.

\bibitem[Sekhari et~al.(2021)Sekhari, Acharya, Kamath, and
  Suresh]{sekhari2021remember}
Ayush Sekhari, Jayadev Acharya, Gautam Kamath, and Ananda~Theertha Suresh.
\newblock Remember what you want to forget: Algorithms for machine unlearning.
\newblock In \emph{Advances in Neural Information Processing Systems},
  volume~34, pages 18075--18086, 2021.

\bibitem[Sharma et~al.(2022)Sharma, Panda, Joshi, and
  Varshney]{sharma2022federated}
Pranay Sharma, Rohan Panda, Gauri Joshi, and Pramod Varshney.
\newblock Federated minimax optimization: Improved convergence analyses and
  algorithms.
\newblock In \emph{International Conference on Machine Learning}, volume 162,
  pages 19683--19730. PMLR, 2022.

\bibitem[Shibata et~al.(2021)Shibata, Irie, Ikami, and
  Mitsuzumi]{shibata2021learning}
Takashi Shibata, Go~Irie, Daiki Ikami, and Yu~Mitsuzumi.
\newblock Learning with selective forgetting.
\newblock In \emph{Proceedings of the Thirtieth International Joint Conference
  on Artificial Intelligence}, pages 989--996. ijcai.org, 2021.

\bibitem[Sinha et~al.(2018)Sinha, Namkoong, and Duchi]{sinhacertifying}
Aman Sinha, Hongseok Namkoong, and John Duchi.
\newblock Certifying some distributional robustness with principled adversarial
  training.
\newblock In \emph{The Sixth International Conference on Learning
  Representations}. OpenReview.net, 2018.

\bibitem[Suriyakumar and Wilson(2022)]{suriyakumar2022algorithms}
Vinith Suriyakumar and Ashia~C Wilson.
\newblock Algorithms that approximate data removal: New results and
  limitations.
\newblock In \emph{Advances in Neural Information Processing Systems},
  volume~35, pages 18892--18903, 2022.

\bibitem[Tarun et~al.(2023)Tarun, Chundawat, Mandal, and
  Kankanhalli]{tarun2023fast}
Ayush~K Tarun, Vikram~S Chundawat, Murari Mandal, and Mohan Kankanhalli.
\newblock Fast yet effective machine unlearning.
\newblock \emph{IEEE Transactions on Neural Networks and Learning Systems},
  2023.

\bibitem[Thekumparampil et~al.(2019)Thekumparampil, Jain, Netrapalli, and
  Oh]{thekumparampil2019efficient}
Kiran~K Thekumparampil, Prateek Jain, Praneeth Netrapalli, and Sewoong Oh.
\newblock Efficient algorithms for smooth minimax optimization.
\newblock In \emph{Advances in Neural Information Processing Systems},
  volume~32, pages 12659--12670, 2019.

\bibitem[Ullah et~al.(2021)Ullah, Mai, Rao, Rossi, and Arora]{ullah2021machine}
Enayat Ullah, Tung Mai, Anup Rao, Ryan~A Rossi, and Raman Arora.
\newblock Machine unlearning via algorithmic stability.
\newblock In \emph{Conference on Learning Theory}, volume 134, pages
  4126--4142. PMLR, 2021.

\bibitem[Vadhan(2017)]{vadhan2017complexity}
Salil Vadhan.
\newblock The complexity of differential privacy.
\newblock \emph{Tutorials on the Foundations of Cryptography: Dedicated to Oded
  Goldreich}, pages 347--450, 2017.

\bibitem[Wang et~al.(2023{\natexlab{a}})Wang, Huai, and Wang]{291009}
Cheng-Long Wang, Mengdi Huai, and Di~Wang.
\newblock Inductive graph unlearning.
\newblock In \emph{32nd USENIX Security Symposium}, pages 3205--3222. USENIX
  Association, 2023{\natexlab{a}}.

\bibitem[Wang et~al.(2023{\natexlab{b}})Wang, Tian, Zhang, Liu, and
  Yu]{wang2023bfu}
Weiqi Wang, Zhiyi Tian, Chenhan Zhang, An~Liu, and Shui Yu.
\newblock Bfu: Bayesian federated unlearning with parameter self-sharing.
\newblock In \emph{Proceedings of the 2023 ACM Asia Conference on Computer and
  Communications Security}, pages 567--578. {ACM}, 2023{\natexlab{b}}.

\bibitem[Warnecke et~al.(2023)Warnecke, Pirch, Wressnegger, and
  Rieck]{warnecke2021machine}
Alexander Warnecke, Lukas Pirch, Christian Wressnegger, and Konrad Rieck.
\newblock Machine unlearning of features and labels.
\newblock In \emph{30th Annual Network and Distributed System Security
  Symposium}. The Internet Society, 2023.

\bibitem[Wu et~al.(2022)Wu, Hashemi, and Srinivasa]{wu2022puma}
Ga~Wu, Masoud Hashemi, and Christopher Srinivasa.
\newblock Puma: Performance unchanged model augmentation for training data
  removal.
\newblock In \emph{Proceedings of the AAAI Conference on Artificial
  Intelligence}, volume~36, pages 8675--8682, 2022.

\bibitem[Wu et~al.(2020)Wu, Dobriban, and Davidson]{wu2020deltagrad}
Yinjun Wu, Edgar Dobriban, and Susan Davidson.
\newblock Deltagrad: Rapid retraining of machine learning models.
\newblock In \emph{International Conference on Machine Learning}, volume 119,
  pages 10355--10366. PMLR, 2020.

\bibitem[Wu et~al.(2023)Wu, Zhu, Li, and He]{wu2023deltaboost}
Zhaomin Wu, Junhui Zhu, Qinbin Li, and Bingsheng He.
\newblock Deltaboost: Gradient boosting decision trees with efficient machine
  unlearning.
\newblock \emph{Proceedings of the ACM on Management of Data}, 1\penalty0
  (2):\penalty0 1--26, 2023.

\bibitem[Xia et~al.(2023)Xia, Liu, Lou, Qin, Ren, Cao, and
  Xiong]{xia2023equitable}
Haocheng Xia, Jinfei Liu, Jian Lou, Zhan Qin, Kui Ren, Yang Cao, and Li~Xiong.
\newblock Equitable data valuation meets the right to be forgotten in model
  markets.
\newblock \emph{Proceedings of the VLDB Endowment}, 16\penalty0 (11):\penalty0
  3349--3362, 2023.

\bibitem[Yan et~al.(2022)Yan, Li, Guo, Li, Li, and Lin]{yan2022arcane}
Haonan Yan, Xiaoguang Li, Ziyao Guo, Hui Li, Fenghua Li, and Xiaodong Lin.
\newblock Arcane: An efficient architecture for exact machine unlearning.
\newblock In \emph{Proceedings of the Thirty-First International Joint
  Conference on Artificial Intelligence}, pages 4006--4013. ijcai.org, 2022.

\bibitem[Yang et~al.(2022)Yang, Hu, Lei, Vashney, Lyu, and
  Ying]{yang2022differentially}
Zhenhuan Yang, Shu Hu, Yunwen Lei, Kush~R Vashney, Siwei Lyu, and Yiming Ying.
\newblock Differentially private sgda for minimax problems.
\newblock In \emph{Uncertainty in Artificial Intelligence}, volume 180, pages
  2192--2202. PMLR, 2022.

\bibitem[Zhang et~al.(2020)Zhang, Wu, Poupart, and Yu]{zhang2020newton}
Guojun Zhang, Kaiwen Wu, Pascal Poupart, and Yaoliang Yu.
\newblock Newton-type methods for minimax optimization.
\newblock \emph{arXiv preprint arXiv:2006.14592}, 2020.

\bibitem[Zhang et~al.(2021)Zhang, Hong, Wang, and
  Zhang]{zhang2021generalization}
Junyu Zhang, Mingyi Hong, Mengdi Wang, and Shuzhong Zhang.
\newblock Generalization bounds for stochastic saddle point problems.
\newblock In \emph{International Conference on Artificial Intelligence and
  Statistics}, volume 130, pages 568--576. PMLR, 2021.

\bibitem[Zhang et~al.(2022{\natexlab{a}})Zhang, Thekumparampil, Oh, and
  He]{zhang2022bring}
Liang Zhang, Kiran~K Thekumparampil, Sewoong Oh, and Niao He.
\newblock Bring your own algorithm for optimal differentially private
  stochastic minimax optimization.
\newblock In \emph{Advances in Neural Information Processing Systems},
  volume~35, pages 35174--35187, 2022{\natexlab{a}}.

\bibitem[Zhang et~al.(2023)Zhang, Lou, Xiong, Zhang, and Liu]{zhang2023closed}
Shuijing Zhang, Jian Lou, Li~Xiong, Xiaoyu Zhang, and Jing Liu.
\newblock Closed-form machine unlearning for matrix factorization.
\newblock In \emph{Proceedings of the 32nd ACM International Conference on
  Information and Knowledge Management}, pages 3278--3287, 2023.

\bibitem[Zhang et~al.(2022{\natexlab{b}})Zhang, Hu, Zhang, and
  He]{zhang2022uniform}
Siqi Zhang, Yifan Hu, Liang Zhang, and Niao He.
\newblock Uniform convergence and generalization for nonconvex stochastic
  minimax problems.
\newblock In \emph{OPT 2022: Optimization for Machine Learning (NeurIPS 2022
  Workshop)}, 2022{\natexlab{b}}.

\bibitem[Zhang et~al.(2022{\natexlab{c}})Zhang, Zhou, Zhao, Che, and
  Lyu]{zhang2022prompt}
Zijie Zhang, Yang Zhou, Xin Zhao, Tianshi Che, and Lingjuan Lyu.
\newblock Prompt certified machine unlearning with randomized gradient
  smoothing and quantization.
\newblock In \emph{Advances in Neural Information Processing Systems},
  volume~35, pages 13433--13455, 2022{\natexlab{c}}.

\end{thebibliography}

\clearpage
\appendix
\setlength\parindent{0pt}
\section{Additional Definitions and Supporting Lemmas}\label{appendix:A}
In this section, we provide additional definitions and supporting lemmas. In the next two sections, Sec.\ref{appendix:B} contains missing proofs in Sec.\ref{sec.scsc} and the online extension to support successive unlearning setting. Sec.\ref{appendix:C} contains missing proofs in Sec.\ref{sec.extentions}, as well as detailed algorithm descriptions for the general convex-concave loss function setting.

\subsection{Additional Definitions}
We first recall the following standard definitions for the loss function $f(\w,\v;z)$ from optimization literature. 
\begin{definition}[\textbf{Function Lipschitz Continuity}]\label{ass:funclip}
    The function $f(\w,\v;z)$ is $L$-Lipschitz, i.e., there exists a constant $L > 0$ such that for all $\w,\w'\in\mathcal W$, $\v,\v'\in\mathcal V$ and $z\in\mathcal Z$, it holds that
    \begin{equation}
        \| f(\w,\v;z) - f(\w',\v';z) \|^2 \leq L^2 (\| \w-\w' \|^2 + \| \v-\v' \|^2).
    \end{equation}
\end{definition}

\begin{definition}[\textbf{Gradient Lipschitz Continuity}]\label{ass:gralip}
    The function $f(\w,\v;z)$ has $\ell$-Lipschitz gradients, i.e., there exists a constant $\ell > 0$ such that for all $\w,\w'\in\mathcal W$, $\v,\v'\in\mathcal V$ and $z\in\mathcal Z$, it holds that
    \begin{equation}
        \| \nabla f(\w,\v;z) - \nabla f(\w',\v';z) \|^2 \leq \ell^2 (\| \w-\w' \|^2 + \| \v-\v' \|^2),
    \end{equation}
    where recall that $\nabla f(\w,\v;z) = \begin{bmatrix}
				\nabla_{\w} f(\w,\v;z)\\
                    \nabla_{\v} f(\w,\v;z)
			\end{bmatrix}$.
\end{definition}

\begin{definition}[\textbf{Hessian Lipschitz Continuity}]\label{ass:hesslip}
    The function $f(\w,\v;z)$ has $\rho$-Lipschitz Hessian, i.e., there exists a constant $\rho > 0$ such that for all $\w,\w'\in\mathcal W$, $\v,\v'\in\mathcal V$ and $z\in\mathcal Z$, it holds that
    \begin{equation}
        \| \nabla^2 f(\w,\v;z) - \nabla^2 f(\w',\v';z) \|^2 \leq \rho^2 (\| \w-\w' \|^2 + \| \v-\v' \|^2),
    \end{equation}
    where recall that $\nabla^2 f(\w,\v;z) = \begin{bmatrix}
        \partial_{\w\w}f(\w,\v;z)  & \partial_{\w\v}f(\w,\v;z) \\
        \partial_{\v\w}f(\w,\v;z) & \partial_{\v\v}f(\w,\v;z)
    \end{bmatrix}$.
\end{definition}

\begin{definition}[\textbf{Strongly-Convex-Strongly-Concave Objective Function}]\label{ass:SC}
    The function $f(\w,\v;z)$ is $\mu_{\w}$-strongly convex on $\mathcal W$ and $\mu_{\v}$-strongly concave on $\mathcal V$, i.e., there exist constants $\mu_{\w} > 0$ and $\mu_{\v} > 0$ such that for all $\w,\w'\in\mathcal W$, $\v,\v'\in\mathcal V$ and $z\in\mathcal Z$, it holds that
    \begin{equation}
    \left \{
        \begin{array}{lr}
            f(\w,\v;z) \geq f(\w',\v;z) + \langle \nabla_{\w} f(\w',\v;z), \w-\w' \rangle + \frac{\mu_{\w}}{2} \| \w-\w' \|^2, \\
            f(\w,\v;z) \leq f(\w,\v';z) + \langle \nabla_{\v} f(\w,\v';z), \v-\v' \rangle - \frac{\mu_{\v}}{2} \| \v-\v' \|^2.
        \end{array}
    \right .
    \end{equation} 
\end{definition}

\begin{definition}[\textbf{Best Response Auxiliary Functions}]
\label{def.aux.functions}
    We introduce auxiliary functions $\VS(\w)$ and $\VSU(\w)$, given by
\begin{equation}\label{eq:aux}
    \VS(\w) := \argmax_{\v\in\mathcal V} \FS(\w,\v), \qquad
    \VSU(\w) := \argmax_{\v\in\mathcal V} \FU(\w,\v),
\end{equation}
and we have $\vn = \VS(\wn)$ and $\vnu = \VSU(\wnu)$. We can similarly introduce ${\tt W}_S(\v)$ and ${\tt W}_{S^{\setminus}}(\v)$ as
\begin{equation}
    {\tt W}_S(\v):= \argmin_{\w\in\mathcal W} \FS(\w,\v),\qquad
    {\tt W}_{S^{\setminus}}(\v) := \argmin_{\w\in\mathcal W}\FU(\w,\v),
\end{equation}
and we have $\wn = {\tt W}_S(\vn)$ and $\wnu = {\tt W}_{S^{\setminus}}(\vnu)$ by this definition.

In addition, we define the primal function $P(\w) := \max_{\v \in \mathcal V}\FS(\w,\v) = \FS(\w,\VS(\w))$, which has gradient $\nabla P(\w) = \nabla_{\w} \FS(\w,\VS(\w))$ and Hessian $\nabla^2_{\w\w} P(\w) = \td \FS(\w,\VS(\w))$ (i.e., the total Hessian of $\FS$). The dual function, its gradient, and Hessian can be similarly defined, e.g., $D(\v) := \min_{\w \in \mathcal W}\FS(\w,\v) = \FS({\tt W}_{S}(\v),\v)$. 
\end{definition}

\subsection{Supporting Lemmas}
The following lemma provides the distance between $\VS(\wn)$ and $\VSU(\wn)$. Similar  result can be derived for the distance between ${\tt W}_S(\vn)$ and ${\tt W}_{S^{\setminus}}(\vn)$.
\begin{lemma}\label{lemma:vsuwn_vn}
    Under Assumption \ref{ass:1} and Assumption\ref{ass:2}, the variables $\VSU(\wn)$ and $\vn$ (i.e., $\VS(\wn)$) defined in Algorithm \ref{alg:learning} satisfy the following distance bound
    \begin{equation}
        \| \vn-\VSU(\wn) \| \leq \frac{2Lm}{\mu_{\v}(n-m)}.
    \end{equation}
\end{lemma}
\begin{proof}
    We observe that 
    \begin{equation}
    \label{eq:vsuvn_right}
        \begin{split}
            & (n-m)  (\FU(\wn,\VSU(\wn)) - \FU(\wn,\vn)) \\
            = & \sum_{z_i \in S\setminus U} f(\wn,\VSU(\wn);z_i) - \sum_{z_i \in S\setminus U} f(\wn,\vn;z_i) \\
            = & \sum_{z_i \in S} f(\wn,\VSU(\wn);z_i) - \sum_{z_i \in U}f(\wn,\VSU(\wn);z_i) - \bigg( \sum_{z_i \in S} f(\wn,\vn;z_i) - \sum_{z_i \in U}(\wn,\vn;z_i) \bigg) \\
            = & n(\FS(\wn,\VSU(\wn)) - \FS(\wn,\vn)) + \sum_{z_i \in U}f(\wn,\vn;z_i) - \sum_{z_i \in U}f(\wn,\VSU(\wn);z_i) \\
            \stackrel{(i)} {\leq} & \sum_{z_i \in U}f(\wn,\vn;z_i) - \sum_{z_i \in U}f(\wn,\VSU(\wn);z_i) 
            \stackrel{(ii)} {\leq}  mL\| \vn-\VSU(\wn) \|,
        \end{split}
    \end{equation}
    where the inequality ($i$) follows from that $\vn$ is the maximizer of the function $\FS(\w,\v)$, thus $\FS(\wn,\VSU(\wn)) - \FS(\wn,\vn) \leq 0$. The inequality ($ii$) is due to the fact that the function $f$ is $L$-Lipschitz. Also note that the function $\FU(\w,\v)$ is $\mu_{\v}$-strongly concave, thus we have
    \begin{equation}
    \label{eq:vsuvn_left}
        \FU(\wn,\VSU(\wn)) - \FU(\wn,\vn) \geq \frac{\mu_{\v}}{2} \| \vn-\VSU(\wn) \|^2.
    \end{equation}
    Eq.(\ref{eq:vsuvn_right}) and eq.(\ref{eq:vsuvn_left}) together give that
    \begin{equation}
        \frac{\mu_{\v} (n-m)}{2} \| \vn-\VSU(\wn) \|^2 \leq mL\| \vn-\VSU(\wn) \|,
    \end{equation}
    which implies that $\| \vn-\VSU(\wn) \| \leq \frac{2Lm}{\mu_{\v}(n-m)}$. 
\end{proof}

The following lemma provides the distance between $(\wnu,\vnu)$ and $(\wn,\vn)$.
\begin{lemma}\label{lemma:wnu_wn}
    Under Assumption \ref{ass:1} and Assumption\ref{ass:2}, the variables $(\wnu,\vnu)$ defined in eq.(\ref{eq:remain_optimal}) and $(\wn,\vn)$ defined in Algorithm \ref{alg:learning} satisfy the following guarantees
    \begin{equation}
        \| \wnu-\wn \| \leq \frac{2Lm}{\mu_{\w}n}, 
        \qquad and \qquad
        \| \vnu-\vn \| \leq \frac{2Lm}{\mu_{\v}n}.
    \end{equation}
\end{lemma}
\begin{proof}
    We begin with the $\w$-part,
    \begin{equation}
    \label{eq:w_right}
        \begin{split}
            & n [\FS(\wnu,\vnu) - \FS(\wn,\vnu)] \\
            = & \sum_{z_i \in S} f(\wnu,\vnu;z_i) - \sum_{z_i \in S} f(\wn,\vnu;z_i) \\
            = & \sum_{z_i \in S\setminus U} f(\wnu,\vnu;z_i) + \sum_{z_i \in U} f(\wnu,\vnu;z_i) - \sum_{z_i \in S\setminus U} f(\wn,\vnu;z_i) - \sum_{z_i \in U}f(\wn,\vnu;z_i) \\
            = & (n-m) [\FU(\wnu,\vnu) - \FU(\wn,\vnu)] + \sum_{z_i \in U} f(\wnu,\vnu;z_i) - \sum_{z_i \in U}f(\wn,\vnu;z_i) \\
            \stackrel{(i)}{\leq} & \sum_{z_i \in U} f(\wnu,\vnu;z_i) - \sum_{z_i \in U}f(\wn,\vnu;z_i) 
            \stackrel{(ii)}{\leq} mL \| \wnu-\wn \|,
        \end{split}
    \end{equation}
    where the inequality ($i$) holds because $\wnu$ is the minimizer of the function $\FU(\w,\v)$, thus $\FU(\wnu,\vnu) - \FU(\wn,\vnu) \leq 0$, 
    and the inequality ($ii$) follows from the fact that the function $f$ is $L$-Lipschitz. Since the function $\FS(\w,\v)$ is $\mu_{\w}$-strongly convex, we further get
    \begin{equation}
    \label{eq:w_left}
        \FS(\wnu,\vnu) - \FS(\wn,\vnu) \geq \frac{\mu_{\w}}{2} \| \wnu-\wn \|^2.
    \end{equation}
    Eq.(\ref{eq:w_right}) and eq.(\ref{eq:w_left}) together gives that
    \begin{equation}
        \frac{\mu_{\w}n}{2} \| \wnu-\wn \|^2 \leq mL \| \wnu-\wn \|.
    \end{equation}
    Thus, we get $\| \wnu-\wn \| \leq \frac{2Lm}{\mu_{\w}n}$.

    For the $\v$-part, we similarly have
    \begin{equation}
    \label{eq:v_right}
        \begin{split}
            & n[\FS(\wnu,\vn) - \FS(\wnu,\vnu)] \\
            = & \sum_{z_i \in S} f(\wnu,\vn;z_i) - \sum_{z_i \in S} f(\wnu,\vnu;z_i) \\
            = & \sum_{z_i \in S\setminus U} f(\wnu,\vn;z_i) + \sum_{z_i \in U} f(\wnu,\vn;z_i) - \sum_{z_i \in S\setminus U} f(\wnu,\vnu;z_i) - \sum_{z_i \in U}f(\wnu,\vnu;z_i) \\
            = & (n-m) [\FU(\wnu,\vn) - \FU(\wnu,\vnu)] + \sum_{z_i \in U} f(\wnu,\vn;z_i) - \sum_{z_i \in U}f(\wnu,\vnu;z_i) \\
            \stackrel{(i)}{\leq} & \sum_{z_i \in U} f(\wnu,\vn;z_i) - \sum_{z_i \in U}f(\wnu,\vnu;z_i) 
            \stackrel{(ii)}{\leq} mL \| \vn-\vnu \|,
        \end{split}
    \end{equation}
    where the inequality ($i$) follows from that $\vnu$ is the maximizer of the function $\FU(\w,\v)$, thus $\FU(\wnu,\vn) - \FU(\wnu,\vnu) \leq 0$. 
    The inequality ($ii$) is due to the fact that the function $f$ is $L$-Lipschitz. In addition, by the strongly-concave assumption of $\FS(\w,\v)$ is $\mu_{\v}$, we have 
    \begin{equation}
    \label{eq:v_left}
        \FS(\wnu,\vn) - \FS(\wnu,\vnu) \geq \frac{\mu_{\v}}{2} \| \vnu-\vn \|^2.
    \end{equation}
    By eq.(\ref{eq:v_right}) and eq.(\ref{eq:v_left}), we get that
    \begin{equation}
        \frac{\mu_{\v}n}{2} \| \vnu-\vn \|^2 \leq mL \| \vn-\vnu \|.
    \end{equation}
    Thus, we have $\| \vnu-\vn \| \leq \frac{2Lm}{\mu_{\v}n}$.    
\end{proof}

In the following, we recall several lemmas (i.e., Lemma \ref{lemma:VS_L} to Lemma \ref{lemma:learning_PD}) from existing minimax optimization literature for completeness.
\begin{lemma}[{\cite[Lemma 4.3]{lin2020gradient}}]
\label{lemma:VS_L}
    Under Assumption \ref{ass:1} and Assumption\ref{ass:2}, for any $\w\in \mathcal W$, the function $\VS(\w)$ is $(\ell/\mu_{\v})$-Lipschitz.
\end{lemma}
\begin{proof}
    By the optimality condition of the function $\VS(\w)$, we have
    \begin{equation}
    \nonumber
        \begin{aligned}
            \langle \nabla_{\v} \FS(\w_1,\VS(\w_1)), \VS(\w_2) - \VS(\w_1) \rangle \leq 0, \\
            \langle \nabla_{\v} \FS(\w_2,\VS(\w_2)), \VS(\w_1) - \VS(\w_2) \rangle \leq 0.    
        \end{aligned}        
    \end{equation}
    Summing the two inequalities above yields
    \begin{equation}
    \label{eq:VSsum}
        \langle \nabla_{\v} \FS(\w_1,\VS(\w_1)) - \nabla_{\v} \FS(\w_2,\VS(\w_2)), \VS(\w_2) - \VS(\w_1) \rangle \leq 0.
    \end{equation}
    Since the function $\FS(\w,\v)$ is $\mu_{\v}$-strongly concave in $\v$, we have
    \begin{equation}
    \label{eq:VSsc}
        \langle \nabla_{\v} \FS(\w_1,\VS(\w_2)) - \FS(\w_1,\VS(\w_1)), \VS(\w_2) - \VS(\w_1) \rangle + \mu_{\v} \| \VS(\w_2) - \VS(\w_1) \| ^ 2 \leq 0.
    \end{equation}
    By eq.(\ref{eq:VSsum}) and eq.(\ref{eq:VSsc}) with the $\ell$-Lipschitz continuity of $\nabla \FS(\w,\v)$, we further get
    \begin{equation}
    \begin{aligned}
        \mu_{\v} \| \VS(\w_2) - \VS(\w_1) \| ^ 2  & \leq \langle \nabla_{\v} \FS(\w_2,\VS(\w_2)) - \nabla_{\v} \FS(\w_1,\VS(\w_2)), \VS(\w_2) - \VS(\w_1) \rangle \\
         & \leq \ell \| \w_2 - \w_1 \| \cdot \| \VS(\w_2) - \VS(\w_1) \|.
    \end{aligned}
    \end{equation}
    Consequently, we have
    \begin{equation}
        \| \VS(\w_2) - \VS(\w_1) \| \leq \frac{\ell}{\mu_{\v}} \| \w_2 - \w_1 \|.
    \end{equation}
\end{proof}
\begin{remark}
    The above lemma can be similarly derived for ${\tt W}_{S}$ to obtain that the best response auxiliary function ${\tt W}_{S}(\v)$ is $(\ell/\mu_{\w})$-Lipschitz. In the next three lemmas, we focus on the $\w$-part and omit the $\v$-part.
\end{remark}

\begin{lemma}[{\citep[Lemma 3]{luo2022finding}}]
\label{lemma.total.hessian.smooth}
    Denote $\kappa_{\v} = \ell/\mu_{\v}$. Under Assumption \ref{ass:1} and Assumption\ref{ass:2}, for any $\w,\w'\in\mathcal W$, we have
    \begin{equation}
        \| \td \FS(\w,\VS(\w)) - \td \FS(\w',\VS(\w')) \| \leq 4\sqrt{2} \kappa_{\v} ^3 \rho \| \w - \w' \|.
    \end{equation}
\end{lemma}

\begin{lemma}[{\citep[Lemma 1]{nesterov2006cubic}}]
\label{lemma:nesterov}
    Denote $\kappa_{\v} = \ell/\mu_{\v}$. Under Assumption \ref{ass:1} and Assumption\ref{ass:2}, for any $\w,\w'\in\mathcal W$, we have
    \begin{equation}
        \| \nabla_{\w} \FS(\w,\VS(\w)) - \nabla_{\w} \FS(\w',\VS(\w')) - \td \FS(\w')(\w-\w') \| \leq \frac{M}{2} \| \w - \w' \|^2,
    \end{equation}
    where $M = 4\sqrt{2} \kappa_{\v} ^3 \rho$.
\end{lemma}
\begin{proof}
    Recall the definition of the primal function $P(\w) := \max_{\v \in \mathcal V}\FS(\w,\v)$ and its gradient $\nabla P(\w) = \nabla_{\w} \FS(\w,\VS(\w))$. Due to the optimality of $\VS$, we have
    \begin{equation}
        \nabla_{\v} \FS(\w,\VS(\w)) = 0.
    \end{equation}
    By taking the total derivative with respect to $\w$, we get
    \begin{equation}
        \partial_{\v\w} \FS(\w,\VS(\w)) + \partial_{\v\v} \FS(\w,\VS(\w)) \nabla \VS(\w)= 0.
    \end{equation}
    Taking the total derivative of $\w$ again on $\nabla P(\w)$, we further have
    \begin{equation}
    \begin{split}
        \nabla^2 P(\w) = & \partial_{\w\w} \FS(\w,\VS(\w)) + \partial_{\w\v} \FS(\w,\VS(\w)) \nabla \VS(\w) \\
        = & \partial_{\w\w} \FS(\w,\VS(\w)) - \partial_{\w\v}\FS(\w,\VS(\w)) \partial_{\v\v}^{-1}\FS(\w,\VS(\w))\partial_{\v\w} \FS(\w,\VS(\w))\\
        = & \td \FS(\w,\VS(\w)).
    \end{split}
    \end{equation}
    Based on the equality of $\nabla^2 P(\w)$ and $\td \FS(\w,\VS(\w))$ above and  Lemma \ref{lemma.total.hessian.smooth}, we have
    \begin{equation}
        \begin{split}
            & \| \nabla_{\w} \FS(\w,\VS(\w)) - \nabla_{\w} \FS(\w',\VS(\w')) - \td \FS(\w')(\w-\w') \| \\
            = & \| \nabla P(\w) - \nabla P(\w') - \nabla^2 P(\w')(\w-\w') \| \\
            \leq & \frac{M}{2} \| \w - \w' \|^2.
        \end{split}
    \end{equation}
\end{proof}
\begin{lemma}\label{lemma:td_lower}
    Under Assumption \ref{ass:1} and Assumption\ref{ass:2}, for all $\w \in \mathcal W$ and $\v \in \mathcal V$, we have $\|\td f(\w,\v;z)\| \leq \ell + \frac{\ell^2}{\mu_{\v}}$.
\end{lemma}

\begin{proof}
    By the definition of the total Hessian, we have
    \begin{equation}
        \begin{split}
            \|\td f(\w,\v;z)\| = & \| \partial_{\w\w} f(\w,\v;z) - \partial_{\w\v}f(\w,\v;z) \partial_{\v\v}^{-1}f(\w,\v;z) \partial_{\v\w}f(\w,\v;z)  \| \\
            \stackrel{(i)}\leq & \| \partial_{\w\w} f(\w,\v;z) \| + \| \partial_{\w\v}f(\w,\v;z) \partial_{\v\v}^{-1}f(\w,\v;z) \partial_{\v\w}f(\w,\v;z)  \| \\
            \stackrel{(ii)}\leq & \ell + \ell \cdot \mu_{\v}^{-1} \cdot \ell = \ell + \frac{\ell^2}{\mu_{\v}},
        \end{split}
    \end{equation}
    where the inequality ($i$) uses the triangle inequality and the inequality ($ii$) is due to the function $f$ has $\ell$-Lipschitz gradients and $f$ is $\mu_{\v}$-strongly concave in $\v$, thus we have $\| \partial_{\w\w} f(\w,\v;z) \| \leq \ell$, $\| \partial_{\w\v} f(\w,\v;z) \| \leq \ell$, $\| \partial_{\v\w} f(\w,\v;z) \| \leq \ell$ and $\| \partial_{\v\v} f(\w,\v;z) \| \leq \mu_{\v}^{-1}$.
\end{proof}
\begin{lemma}[{\citep[Lemma 4.4]{zhang2022bring}}]
\label{lemma:learning_PD}
    Under Assumption \ref{ass:1} and Assumption\ref{ass:2}, the population weak PD risk for the minimax learning variables $(\wn,\vn)$ returned by Algorithm \ref{alg:learning} has
    \begin{equation}
        \triangle^w(\wn,\vn) \leq \frac{2\sqrt{2} L^2}{\mu n}.
    \end{equation}
\end{lemma}
\begin{lemma}[{\citep[Theorem 2]{zhang2021generalization}}]
\label{lemma:learning_strPD}
    Under Assumption \ref{ass:1} and Assumption\ref{ass:2}, the population strong PD risk for the minimax learning variables $(\wn,\vn)$ returned by Algorithm \ref{alg:learning} has
    \begin{equation}
        \triangle^s(\wn,\vn) \leq \frac{2\sqrt{2} L^2}{n}\cdot\sqrt{\frac{\ell^2}{\mu_{\w}\mu_{\v}} + 1}\cdot\left(\frac{1}{\mu_{\w}} + \frac{1}{\mu_{\v}}\right) \leq \frac{8L^2 \ell}{\mu^2 n}.
    \end{equation}
\end{lemma}
\section{Detailed Algorithm Analysis and Missing Proofs in Section 4}
\label{appendix:B}
\subsection{Analysis for Algorithm \ref{alg:unlearning}}
\label{appendix:B.0}
In the following, we provide the analysis for Algorithm \ref{alg:unlearning} in terms of guarantees of $(\epsilon,\delta)$-certified unlearning, population primal-dual risk, and deletion capacity and the corresponding proofs.
\begin{lemma}[\textbf{Closeness Upper Bound}]
\label{lem:sensitivity}
    Suppose the loss function $f$ satisfies Assumption \ref{ass:1} and \ref{ass:2}, $\|\td \FS(\wn,\vn)\| \geq \mu_{\w\w}$ and $\| \tdv \FS(\wn,\vn) \| \geq \mu_{\v\v}$. Let $\mu = \min\{\mu_{\w},\mu_{\v},\mu_{\w\w},\mu_{\v\v}\}$. Then, we have the closeness bound between $(\widehat{\w},\widehat{\v})$ in Line 2 of Algorithm \ref{alg:unlearning} and $(\wnu,\vnu)$ in eq.(\ref{eq:remain_optimal}):
    \begin{equation}
       \{ \|\wnu-\hw\|, \|\vnu-\hv\|\} \leq \frac{(8 \sqrt{2}L^2 \ell^3 \rho / \mu^5 + 8 L \ell^2 / \mu^2) m^2}{n ( \mu n - (\ell + \ell^2/\mu)m )}.
    \end{equation}
\end{lemma}

\begin{proof}
    Recall that the empirical loss functions $\FU(\w,\v)$ and $\FS(\w,\v)$ are
    \begin{equation}
        \FU(\w,\v) := \frac{1}{n-m}\sum_{z_i\in S\setminus U} f(\w,\v;z_i), \quad \text{and} \quad
        \FS(\w,\v) := \frac{1}{n} \sum_{z_i \in S} f(\w,\v;z_i).
    \end{equation}
    We focus on the key term $\nabla_{\w} \FU(\wnu,\VS(\wnu)) -  \nabla_{\w} \FU(\wn,\vn) - \td \FU(\wn,\vn)(\wnu - \wn)$, which has the following conversions
    \begin{equation}
    \label{eq:maineq_left}
        \begin{split}
          &  \|\nabla_{\w} \FU(\wnu,\VS(\wnu)) -  \nabla_{\w} \FU(\wn,\vn) - \td \FU(\wn,\vn)(\wnu - \wn)\| \\
        = & \|\frac{n}{n-m} [\nabla_{\w} \FS(\wnu,\VS(\wnu)) -  \nabla_{\w} \FS(\wn,\vn) - \td \FS(\wn,\vn)(\wnu - \wn)] \\
          & - \frac{1}{n-m} \sum_{z_i \in U} [\nabla_{\w} f(\wnu,\VS(\wnu);z_i) - \nabla_{\w} f(\wn,\vn;z_i)] \\
          & + \frac{1}{n-m} \sum_{z_i \in U} \td f(\wn,\vn;z_i)(\wnu - \wn) \| \\
     \leq & \frac{n}{n-m} \| \nabla_{\w} \FS(\wnu,\VS(\wnu)) -  \nabla_{\w} \FS(\wn,\vn) - \td \FS(\wn,\vn)(\wnu - \wn) \| \\
          & + \frac{1}{n-m} \sum_{z_i \in U} \|\nabla_{\w} f(\wnu,\VS(\wnu);z_i) - \nabla_{\w} f(\wn,\vn;z_i)\| \\
          & + \frac{1}{n-m} \| \sum_{z_i \in U} \td f(\wn,\vn;z_i)(\wnu - \wn) \|.
        \end{split} 
    \end{equation}
    We denote $\kappa_{\v} = \ell/\mu_{\v}$. For the first term on the right-hand side of the inequality in eq.(\ref{eq:maineq_left}), we have
    \begin{equation}
    \label{eq:right_1}
        \begin{split}
            & \frac{n}{n-m} \| \nabla_{\w} \FS(\wnu,\VS(\wnu)) -  \nabla_{\w} \FS(\wn,\vn) - \td \FS(\wn,\vn)(\wnu - \wn) \| \\
        \stackrel{(i)}\leq& \frac{n}{n-m} \cdot 2\sqrt{2} \kappa_{\v}^3 \rho \| \wnu - \wn \|^2 \stackrel{(ii)}\leq \frac{8\sqrt{2} \kappa_{\v}^3 \rho L^2 m^2}{\mu_{\w}^2 n(n-m)}
        \leq \frac{8\sqrt{2}\rho L^2 \ell^3 m^2}{\mu^5 n(n-m)},
        \end{split}
    \end{equation}
    where the inequality ($i$) is by  Lemma \ref{lemma:nesterov} and the inequality ($ii$) is by Lemma \ref{lemma:wnu_wn}.
    
    For the second term on the right-hand side of the inequality in eq.(\ref{eq:maineq_left}), we have
    \begin{equation}
    \label{eq:right_3}
        \begin{split}
            & \frac{1}{n-m} \sum_{z_i \in U} \| \nabla_{\w} f(\wnu,\VS(\wnu);z_i) - \nabla_{\w} f(\wn,\vn;z_i) \| \\
        \stackrel{(i)}\leq& \frac{1}{n-m} \cdot m\ell \sqrt{\| \wnu - \wn \|^2 + \| \VS(\wnu) - \VS(\wn) \|^2} \\
        \stackrel{(ii)}\leq& \frac{1}{n-m} \cdot m\ell \sqrt{\| \wnu - \wn \|^2 + \kappa_{\v}^2\| \wnu - \wn \|^2} \\
        \stackrel{(iii)}\leq& 
        \frac{2Llm^2 \sqrt{1 + \kappa_{\v}^2} }{\mu_{\w}n(n-m)} \leq \frac{2 \sqrt{2} L l\kappa_{\v} m^2}{\mu n(n-m)}
        \leq \frac{2 \sqrt{2} L l^2 m^2}{\mu^2 n(n-m)},
        \end{split}
    \end{equation}
    where the inequality ($i$) follows by the fact that the function $\nabla_{\w} f(\cdot,\cdot)$ is $\ell$-Lipschitz continuous and $\vn = \VS(\wn)$. The inequality ($ii$) holds because Lemma \ref{lemma:VS_L}, and the inequality ($iii$) is by Lemma \ref{lemma:wnu_wn}.

    For the third term on the right-hand side of the inequality in eq.(\ref{eq:maineq_left}), we have
    \begin{equation}
    \label{eq:right_4}
        \begin{split}
            &\frac{1}{n-m} \| \sum_{z_i \in U} \td f(\wn,\vn;z_i)(\wnu - \wn) \| \\
            \leq & (\ell + \frac{\ell^2}{\mu_{\v}}) \cdot \frac{2L m^2}{\mu_{\w} n(n-m)} 
            \leq \frac{4L\ell^2 m^2}{\mu^2 n(n-m)},       
        \end{split}
    \end{equation}
    where the first inequality is by Lemma \ref{lemma:td_lower}. Plugging eq.(\ref{eq:right_1}), eq.(\ref{eq:right_3}) and eq.(\ref{eq:right_4}) into eq.(\ref{eq:maineq_left}), we further get
    \begin{equation}
    \label{eq:maineq_left_all}
        \begin{split}
            & \|\nabla_{\w} \FU(\wnu,\VS(\wnu)) -  \nabla_{\w} \FU(\wn,\vn) - \td \FU(\wn,\vn)(\wnu - \wn)\| \\
            \leq & (8 \sqrt{2}L^2 \ell^3 \rho / \mu^5 + 8 L \ell^2 / \mu^2) \frac{m^2}{n(n-m)}.
        \end{split}
    \end{equation}
    The above derivation yields an upper bound result. In the following, we derive a lower bound result. Let $\x$ be the vector satisfying the following relation,
    \begin{equation} \label{eq:vec_x}
        \wnu = \wn + \frac{1}{n-m} [\td \FU(\wn,\vn)]^{-1} \sum_{z_i\in U} \nabla_{\w} f(\wn,\vn;z_i) + \x.
    \end{equation}
    Since we have $\nabla_{\w} \FU(\wn,\vn) = -\frac{1}{n-m} \sum_{z_i\in U} \nabla_{\w} f(\wn,\vn;z_i)$ and $\nabla_{\w} \FU(\wnu,\VS(\wnu)) = 0$ due to the optimality of $\wnu$, plugging eq.(\ref{eq:vec_x}) into eq.(\ref{eq:maineq_left_all}), we get that 
    \begin{equation}\label{eq:td_x}
        \| \td \FU(\wn,\vn) \cdot \x\| \leq (8 \sqrt{2}L^2 \ell^3 \rho / \mu^5 + 8 L \ell^2 / \mu^2) \frac{m^2}{n(n-m)}.
    \end{equation}
    For the left-hand side of eq.(\ref{eq:td_x}), with $\| \td \FS(\wn,\vn) \| \geq \mu_{\w\w}$, we have
    \begin{equation}
    \label{eq:maineq_right}
        \begin{split}
            \| \td \FU(\wn,\vn) \cdot \x\| & = \frac{1}{n-m} \| [\sum _{z_i \in S\setminus U}\td f(\wn,\vn;z_i)] \cdot \x \| \\
            & = \frac{1}{n-m} \| [\sum _{z_i \in S}\td f(\wn,\vn;z_i) - \sum _{z_i \in U}\td f(\wn,\vn;z_i)] \cdot \x \| \\
            & \geq \frac{1}{n-m} \bigg(\| n \td \FS(\wn,\vn) \| - \| \sum _{z_i \in U}\td f(\wn,\vn;z_i) \|\bigg) \cdot \| \x\| \\
            & \geq \frac{(\mu_{\w\w}n - (\ell + \ell^2/\mu_{\v})m)}{n-m}\| \x \| \geq \frac{(\mu n - (\ell + \ell^2/\mu)m)}{n-m}\| \x \|,
        \end{split}
    \end{equation}
    where the second inequality is by Lemma \ref{lemma:td_lower}. Combining eq.(\ref{eq:td_x}), eq.(\ref{eq:maineq_left_all}), and the definition of the vector $\| \x \|$, we get that
    \begin{equation}
        \|\wnu-\hw\| = \| \x \| \leq \frac{(8 \sqrt{2}L^2 \ell^3 \rho / \mu^5 + 8 L \ell^2 / \mu^2) m^2}{n ( \mu n - (\ell + \ell^2/\mu)m )}.
    \end{equation}
    Symmetrically, we can get that $\| \vnu-\hv \| \leq \frac{(8 \sqrt{2}L^2 \ell^3 \rho / \mu^5 + 8 L \ell^2 / \mu^2) m^2}{n ( \mu n - (\ell + \ell^2/\mu)m )}$.   
\end{proof}

\begin{theorem} [\textbf{$(\epsilon,\delta)$-Minimax Unlearning Certification}]
\label{theorem:certi}
    Under the same settings of Lemma \ref{lem:sensitivity}, our minimax learning algorithm $A_{sc-sc}$ and unlearning algorithm $\bar A_{sc-sc}$ is $(\epsilon,\delta)$-certified minimax unlearning if we choose
    \begin{equation}
        \sigma_1 \text{~and~} \sigma_2 = \frac{2(8 \sqrt{2}L^2 \ell^3 \rho / \mu^5 + 8 L \ell^2 / \mu^2) m^2}{n ( \mu n - (\ell + \ell^2/\mu)m ) \epsilon} \sqrt{2\log(2.5/\delta)}.
    \end{equation}
\end{theorem}

\begin{proof}
Our proof for $(\epsilon,\delta)$-minimax unlearning certification is similar to the one used for the differential privacy guarantee of the Gaussian mechanism \citep{dwork2014algorithmic}.

Let $(\wn,\vn)$ be the output of the learning algorithm $A_{sc-sc}$ trained on dataset $S$ and $(\wu,\vu)$ be the output of the unlearning algorithm $\bar A_{sc-sc}$ running with delete requests $U$, the learned model $(\wn,\vn)$, and the memory variables $T(S)$. Then we have $(\wn,\vn) = A_{sc-sc}(S)$ and $(\wu,\vu) = \bar A_{sc-sc}(U, A_{sc-sc}(S), T(S))$. We also denote the intermediate variables before adding noise in algorithm $\bar A_{sc-sc}$ as $(\hw,\hv)$, and we have $\wu = \hw + \bm\xi_1$ and $\vu = \hv + \bm\xi_2$.

Smilarly, let $(\wnu,\vnu)$ be the output of the learning algorithm $A_{sc-sc}$ trained on dataset $S\setminus U$ and $(\wuu,\vuu)$ be the output of the unlearning algorithm $\bar A_{sc-sc}$ running with delete requests $\emptyset$, the learned model $(\wnu,\vnu)$, and the memory variables $T(S\setminus U)$. Then we have $(\wnu,\vnu) = A_{sc-sc}(S\setminus U)$ and $(\wuu,\vuu) = \bar A_{sc-sc}(\emptyset, A_{sc-sc}(S\setminus U), T(S))$. We also denote the intermediate variables before adding noise in algorithm $\bar A_{sc-sc}$ as $(\hwu,\hvu)$, and we have $\wuu = \hwu + \bm\xi_1$ and $\vuu = \hvu + \bm\xi_2$. Note that $\hwu=\wnu$ and $\hvu=\vnu$.

We sample the noise $\bm{\xi}_1\sim \mathcal N(0,\sigma_1\mathbf{I}_{d_1})$ and $\bm{\xi}_2\sim \mathcal N(0,\sigma_2\mathbf{I}_{d_2})$ with the scale:
\begin{equation}
    \left\{
        \begin{array}{lr}
             \sigma_1 = \|\wnu-\hw\| \cdot \frac{\sqrt{2\log(2.5/\delta)}}{\epsilon/2} = \|\hwu-\hw\| \cdot\frac{\sqrt{2\log(2.5/\delta)}}{\epsilon/2},  \\
             \sigma_2= \|\vnu-\hv\| \cdot \frac{\sqrt{2\log(2.5/\delta)}}{\epsilon/2} = \|\hvu-\hv\| \cdot\frac{\sqrt{2\log(2.5/\delta)}}{\epsilon/2},
        \end{array}
    \right .
\end{equation}
where $\|\wnu-\hw\|$ and $\|\wnu-\hw\|$ are given in Lemma \ref{lem:sensitivity}. Then, following the same proof as \citet[Theorem A.1]{dwork2014algorithmic} together with the composition property of DP \citep[Lemma 7.2.3]{vadhan2017complexity}, we get that, for any set $O\subseteq \varTheta$ where $\varTheta:=\mathcal W\times \mathcal V$,
\begin{equation}
    \Pr[(\wu,\vu)\in O] \leq e^\epsilon \Pr[(\wuu,\vuu) \in O] + \delta,\; \text{and} \;
    \Pr[(\wuu,\vuu)\in O] \leq e^\epsilon \Pr[(\wu,\vu) \in O] + \delta,
\end{equation}
which implies that the algorithm pair $A_{sc-sc}$ and $\bar A_{sc-sc}$ is $(\epsilon,\delta)$-certified minimax unlearning.
\end{proof}

\begin{theorem}[\textbf{Population Primal-Dual Risk}]
\label{theorem:guarantee_SC}
    Under the same settings of Lemma \ref{lem:sensitivity} and denote $d=\max\{d_1,d_2\}$, the population weak and strong PD risk for the certified minimax unlearning variables $(\wu,\vu)$ returned by Algorithm \ref{alg:unlearning} are
        \begin{equation}
            \left \{
                \begin{aligned}
                    & \triangle^w (\wu,\vu) = \mathcal O \left( (L^3 \ell^3 \rho / \mu^6 + L^2 \ell^2/\mu^3)\cdot \frac{m^2 \sqrt{d\log(1/\delta)}}{n^2 \epsilon} + \frac{mL^2}{\mu n}\right), \\
                    & \triangle^s (\wu,\vu) = \mathcal O \left( (L^3 \ell^3 \rho / \mu^6 + L^2 \ell^2/\mu^3)\cdot \frac{m^2 \sqrt{d\log(1/\delta)}}{n^2 \epsilon} + \frac{mL^2}{\mu n} + \frac{L^2\ell}{\mu^2 n}\right).
                \end{aligned}
            \right .
        \end{equation}
\end{theorem}

\begin{proof}
    We begin with the population weak PD risk for the certified minimax unlearning variable $(\wu, \vu)$, which has the following conversions,
    \begin{equation}\label{eq:unlearning_PD}
        \begin{aligned}
            & \triangle ^w(\wu,\vu) \\
            = & \max_{\v\in\mathcal V}\mathbb E[F(\wu,\v)]-\min_{\w\in\mathcal W}\mathbb E[F(\w,\vu)] \\
            = & \max_{\v\in\mathcal V}\mathbb E[F(\wu,\v) - F(\wn,\v) + F(\wn,\v)] 
               - \min_{\w\in\mathcal W}\mathbb E[F(\w,\vu) - F(\w,\vn) + F(\w,\vn)] \\
            \leq & \max_{\v\in\mathcal V}\mathbb E[F(\wu,\v) - F(\wn,\v)] + \max_{\v\in\mathcal V} \mathbb E[F(\wn,\v)] 
             - \min_{\w\in\mathcal W}\mathbb E[F(\w,\vu) - F(\w,\vn)] - \min_{\w\in\mathcal W}\mathbb E[F(\w,\vn)]\\
            = & \max_{\v\in\mathcal V}\mathbb E[F(\wu,\v) - F(\wn,\v)] + \max_{\w\in\mathcal W} \mathbb E[(-F)(\w,\vu) - (-F)(\w,\vn)] \\
              & + \max_{\v\in\mathcal V} \mathbb E[F(\wn,\v)] - \min_{\w\in\mathcal W}\mathbb E[F(\w,\vn)]\\
            \stackrel{(i)}\leq & \mathbb E[L \|\wu-\wn\|] + \mathbb E[L \|\vu-\vn\|] + \triangle^w(\wn,\vn) \\
            \stackrel{(ii)}\leq & \mathbb E[L \|\wu-\wn\|] + \mathbb E[L \|\vu-\vn\|] + \frac{2\sqrt{2}L^2}{\mu n},
        \end{aligned}
    \end{equation}
    where the inequality ($i$) holds because the population loss function $F(\w,\v):=\mathbb E[f(\w,\v;z)]$
    is $L$-Lipschitz continuous. The inequality ($ii$) is by Lemma \ref{lemma:learning_PD}.
    
    By recalling the unlearning update step in Algorithm \ref{alg:unlearning}, we have
    \begin{equation}
    \label{eq:wu}
        \wu = \wn+\frac{1}{n-m}[\td \FU(\wn,\vn)]^{-1}\sum_{z_i\in U}\nabla_{\w} f(\wn,\vn;z_i)+\bm{\xi}_1,
    \end{equation}
    where the vector $\bm{\xi}_1\in\mathbb R^{d_1}$ is drawn independently from $\mathcal{N}(0,\sigma_1^2 \mathbf{I}_{d_1})$. From the relation in eq.(\ref{eq:wu}), we further get
    \begin{equation}\label{eq:wu_wn}
        \begin{aligned}
            \mathbb E[\|\wu-\wn\|]=&\mathbb E\left[\left\|\frac{1}{n-m}[\td \FU(\wn,\vn)]^{-1}\cdot \sum_{z_i\in U} \nabla_{\w} f(\wn,\vn;z_i) + \bm{\xi}_1\right\|\right]\\
            \stackrel{(i)}\leq&\frac{1}{n-m}\mathbb E\left[\left\|[\td \FU(\wn,\vn)]^{-1}\cdot \sum_{z_i\in U} \nabla_{\w} f(\wn,\vn;z_i)\right\|\right]+\mathbb E[\|\bm{\xi}_1\|]\\
            \stackrel{(ii)}\leq&\frac{1}{n-m}\cdot\frac{n-m}{(\mu n - \ell (1+\ell/\mu)m )} \mathbb E\left[\left\|\sum_{z_i\in U} \nabla_{\w} f(\wn,\vn;z_i)\right\|\right]+\sqrt{\mathbb E[\|\bm{\xi}_1\|^2]}\\
            \stackrel{(iii)}=&\frac{1}{(\mu n - \ell (1+\ell/\mu)m )} \mathbb E\left[\left\|\sum_{z_i\in U} \nabla_{\w} f(\wn,\vn;z_i)\right\|\right]+\sqrt{d_1}\sigma_1\\
            \stackrel{(iv)}\leq&\frac{mL}{(\mu n - \ell (1+\ell/\mu)m )}+\sqrt{d_1}\sigma_1,
        \end{aligned}
    \end{equation}
    where the inequality ($i$) is by the triangle inequality and the inequality ($ii$) follows from the relation in eq.(\ref{eq:maineq_right}), together with the Jensen's inequality to bound $\mathbb E[\|\bm{\xi}_1\|]$. The equality ($iii$) holds because the vector $\bm{\xi}_1\sim\mathcal{N}(0,\sigma_1^2 \mathbf{I}_{d_1})$ and thus we have $\mathbb E[\|\bm{\xi}_1\|^2]=d_1\sigma_1^2$. Furthermore, the inequality ($iv$) is due to the fact that $f(\w,\v;z)$ is $L$-Lipshitz continuous. 
    
    Symmetrically, we have
    \begin{equation}\label{eq:vu_vn}
        \mathbb E[\|\vu-\vn\|] \leq \frac{mL}{(\mu n - \ell (1+\ell/\mu)m )}+\sqrt{d_2}\sigma_2.
    \end{equation}
    Plugging eq.(\ref{eq:wu_wn}) and eq.(\ref{eq:vu_vn}) into eq.(\ref{eq:unlearning_PD}) with $d=\max\{d_1,d_2\}$ we get
    \begin{equation}
        \triangle ^w (\wu,\vu) \leq \frac{2mL^2}{(\mu n - \ell (1+\ell/\mu)m )} + \sqrt{d}(\sigma_1 + \sigma_2)L + \frac{2\sqrt{2}L^2}{\mu n}.
    \end{equation}
    With the noise scale $\sigma_1 $ and $\sigma_2 $  being equal to $ \frac{2(8 \sqrt{2}L^2 \ell^3 \rho / \mu^5 + 8 L \ell^2 / \mu^2) m^2}{n ( \mu n - (\ell + \ell^2/\mu)m ) \epsilon} \sqrt{2\log(2.5/\delta)}$, we can get our generalization guarantee with population weak PD risk:
    \begin{equation}
        \triangle^w (\wu,\vu) = \mathcal O \left( (L^3 \ell^3 \rho / \mu^6 + L^2 \ell^2/\mu^3)\cdot \frac{m^2 \sqrt{d\log(1/\delta)}}{n^2 \epsilon} + \frac{mL^2}{\mu n}\right).
    \end{equation}
    For the population strong PD risk $\triangle ^s (\wu,\vu)$, similarly, we have
    \begin{equation}\label{eq:unlearning_strPD}
        \begin{aligned}
            & \mathbb E[\max_{\v\in\mathcal V}F(\wu,\v)-\min_{\w\in\mathcal W}F(\w,\vu)] \\
            = & \mathbb E[\max_{\v\in\mathcal V}\left(F(\wu,\v) - F(\wn,\v) + F(\wn,\v)\right)
               - \min_{\w\in\mathcal W}\left(F(\w,\vu) - F(\w,\vn) + F(\w,\vn)\right)] \\
            = & \mathbb E[\max_{\v\in\mathcal V}(F(\wu,\v) - F(\wn,\v) + F(\wn,\v))]
                - \mathbb E[\min_{\w\in\mathcal W}(F(\w,\vu) - F(\w,\vn) + F(\w,\vn))] \\
            \leq &  \mathbb E[\max_{\v\in\mathcal V}(F(\wu,\v) - F(\wn,\v)) + \max_{\v\in\mathcal V}F(\wn,\v)] 
                - \mathbb E[\min_{\w\in\mathcal W}(F(\w,\vu) - F(\w,\vn)) + \min_{\w\in\mathcal W}F(\w,\vn)] \\
            = & \mathbb E[\max_{\v\in\mathcal V}(F(\wu,\v) - F(\wn,\v))] + \mathbb E[\max_{\v\in\mathcal V}F(\wn,\v)]  
            - \mathbb E[\min_{\w\in\mathcal W}(F(\w,\vu) - F(\w,\vn))] -\mathbb E[\min_{\w\in\mathcal W}F(\w,\vn)] \\
            = & \mathbb E[\max_{\v\in\mathcal V}(F(\wu,\v) - F(\wn,\v))] + \mathbb E[\max_{\w\in\mathcal W}((-F)(\w,\vu) - (-F)(\w,\vn))] \\ 
            & +\mathbb E[\max_{\v\in\mathcal V}F(\wn,\v) - \min_{\w\in\mathcal W}F(\w,\vn)]\\
            \stackrel{(i)}\leq & \mathbb E[L \|\wu-\wn\|] + \mathbb E[L \|\vu-\vn\|] + \triangle^s(\wn,\vn)\\
            \stackrel{(ii)}\leq & \frac{2mL^2}{(\mu n - \ell (1+\ell/\mu)m )} + \sqrt{d}(\sigma_1 + \sigma_2)L + \frac{8L^2\ell}{\mu^2 n},
        \end{aligned}
    \end{equation}
    where inequality ($i$) is due to the fact that the population loss function $F(\w,\v):=\mathbb E[f(\w,\v;z)]$ is $L$-Lipschitz continuous. The inequality ($ii$) uses eq.(\ref{eq:wu_wn}), eq.(\ref{eq:vu_vn}) and Lemma \ref{lemma:learning_strPD}. With the same noise scale above, we can get the generalization guarantee in terms of strong PD risk below,
    \begin{equation}
        \triangle^s (\wu,\vu) = \mathcal O \left( (L^3 \ell^3 \rho / \mu^6 + L^2 \ell^2/\mu^3)\cdot \frac{m^2 \sqrt{d\log(1/\delta)}}{n^2 \epsilon} + \frac{mL^2}{\mu n} + \frac{L^2\ell}{\mu^2 n}\right).
    \end{equation}
\end{proof}

\begin{theorem}[\textbf{Deletion Capacity}]
\label{theorem:del_cap}
    Under the same settings of Lemma \ref{lem:sensitivity} and denote $d=\max\{d_1,d_2\}$, the deletion capacity of Algorithm \ref{alg:unlearning} is
    \begin{equation}
        m ^ { A,\bar A} _ {\epsilon,\delta,\gamma}(d_1,d_2,n) \geq c \cdot \frac{n\sqrt{\epsilon}}{(d\log(1/\delta))^{1/4}},
    \end{equation}
    where the constant $c$ depends on $L, l, \rho,$ and $\mu$ of the loss function $f$.
\end{theorem}

\begin{proof}
    By the definition of deletion capacity, in order to ensure the population PD risk derived in Theorem \ref{theorem:guarantee_SC} is bounded by $\gamma$, it suffices to let:
\begin{equation}\nonumber
    m ^ { A,\bar A} _ {\epsilon,\delta,\gamma}(d_1,d_2,n) \geq c \cdot \frac{n\sqrt{\epsilon}}{(d\log(1/\delta))^{1/4}},
\end{equation}
where the constant $c$ depends on the properties of the loss function $f(\w,\v;z)$.
\end{proof}

\subsection{Missing Proofs of Sec.\ref{subsec.analysis.eff}}
\subsubsection{Proof of Lemma \ref{lem:sensitivity.eff} (Closeness Upper Bound)}
\begin{proof}
    By the definition of the functions $\FS(\w,\v)$ and $\FU(\w,\v)$, we have
    \begin{equation}\label{eq:maineq_eff}
        \begin{split}
            & \| \nabla_{\w}\FU(\wnu,\VS(\wnu)) - \nabla_{\w}\FU(\wn,\vn) - \frac{n}{n-m}\td \FS(\wn,\vn)(\wnu-\wn))\| \\
            = & \| \frac{n}{n-m}[\nabla_{\w}\FS(\wnu,\VS(\wnu)) - \nabla_{\w}\FS(\wn,\vn) - \td\FS(\wn,\vn)(\wnu-\wn)] \\
            & - \frac{1}{n-m}\sum_{z_i\in U}[\nabla_{\w} f(\wnu,\VS(\wnu);z_i) - \nabla_{\w} f(\wn,\vn;z_i)] \|\\
            \stackrel{(i)}\leq & \frac{n}{n-m} \| \nabla_{\w}\FS(\wnu,\VS(\wnu)) - \nabla_{\w}\FS(\wn,\vn) - \td\FS(\wn,\vn)(\wnu-\wn) \| \\
            & + \frac{1}{n-m} \sum_{z_i\in U}\| \nabla_{\w} f(\wnu,\VS(\wnu);z_i) - \nabla_{\w} f(\wn,\vn;z_i) \| \\
            \stackrel{(ii)}\leq & \frac{8\sqrt{2}\rho L^2 \ell^3 m^2}{\mu^5 n(n-m)} + \frac{2 \sqrt{2} L l^2 m^2}{\mu^2 n(n-m)},
        \end{split}
    \end{equation}
    where the inequality ($i$) holds because the triangle inequality and the inequality ($ii$) uses the results in eq.(\ref{eq:right_1}) and eq.(\ref{eq:right_3}). Now let $\widetilde {\x}$ be the vector satisfying the following relation,
    \begin{equation}
        \wnu = \wn + \frac{1}{n}[\td \FS(\wn,\vn)]^{-1} \sum_{z_i\in U}\nabla_{\w}f(\wn,\vn;z_i) + \widetilde {\x}.
    \end{equation}
    Since we have $\nabla_{\w} \FU(\wn,\vn) = -\frac{1}{n-m} \sum_{z_i\in U} \nabla_{\w} f(\wn,\vn;z_i)$ and $\nabla_{\w} \FU(\wnu,\VS(\wnu)) = 0$ due to the optimality of $\wnu$, plugging the above relation into eq.(\ref{eq:maineq_eff}), we get
    \begin{equation}\label{eq:eff_x_right}
        \|\frac{n}{n-m} \td \FS(\wn,\vn) \widetilde{\x}\| \leq (8 \sqrt{2}L^2 \ell^3 \rho / \mu^5 + 2 \sqrt{2} L \ell^2 / \mu^2) \frac{m^2}{n(n-m)}.
    \end{equation}
    With $\td \FS(\wn,\vn) \geq \mu_{\w\w}$, we also have
    \begin{equation}\label{eq:eff_x_left}
        \|\frac{n}{n-m} \td \FS(\wn,\vn) \widetilde{\x}\| \geq \frac{\mu_{\w\w}n}{n-m}\| \widetilde{\x}\| \geq \frac{\mu n}{n-m}\| \widetilde{\x}\|.
    \end{equation}
    Combining eq.(\ref{eq:eff_x_left}), eq.(\ref{eq:eff_x_right}), and the definition of the vector $\| \widetilde{\x}\|$, we get that
    \begin{equation}
        \|\wnu-\tw\| = \| \widetilde{\x}\| \leq \frac{(8\sqrt{2}L^2 \ell^3 \rho / \mu^6 + 2\sqrt{2}L \ell^2 / \mu^3) m^2}{n^2}.
    \end{equation}
    Symmetrically, we can get $\|\vnu - \tv\| \leq \frac{(8\sqrt{2}L^2 \ell^3 \rho / \mu^6 + 2\sqrt{2}L \ell^2 / \mu^3) m^2}{n^2}$.
\end{proof}

\subsubsection{Proof of Theorem \ref{theorem:certi.eff} ($(\epsilon,\delta)$-Minimax Unlearning Certification)}
\begin{proof}
    With the closeness upper bound in Lemma \ref{lem:sensitivity.eff} and the given noise scales in eq.(\ref{eq:noise.eff_}), the proof is identical to that of Theorem \ref{theorem:certi}.
\end{proof}

\subsubsection{Proof of Theorem \ref{theorem:guarantee_SC.eff} (Population Primal-Dual Risk)}
\begin{proof}
    We start with the population weak PD risk. By eq.(\ref{eq:unlearning_PD}), we have
    \begin{equation}\label{eq:eff_PD}
        \triangle^w(\tw^u,\tv^u) \leq \mathbb E[L \|\tw^u-\wn\|] + \mathbb E[L \|\tv^u-\vn\|] + \frac{2\sqrt{2}L^2}{\mu n}.
    \end{equation}
    By recalling the unlearning step in Algorithm \ref{alg:unlearning.extension.online}, we have 
    \begin{equation}
        \tw^u = \wn + \frac{1}{n}[\td \FS(\wn,\vn)]^{-1} \sum_{z_i \in U}\nabla_{\w}f(\wn,\vn;z_i) + \bm{\xi}_1,
    \end{equation}
    where the vector $\bm{\xi}_1\in\mathbb R^{d_1}$ is drawn independently from $\mathcal{N}(0,\sigma_1^2 \mathbf{I}_{d_1})$. From the relation in eq.(\ref{eq:wu}), we further get
    \begin{equation}\label{eq:eff_wu_wn}
        \begin{aligned}
            \mathbb E[\|\tw^u-\wn\|]=&\mathbb E\left[\left\|\frac{1}{n}[\td \FS(\wn,\vn)]^{-1}\cdot \sum_{z_i\in U} \nabla_{\w} f(\wn,\vn;z_i) + \bm{\xi}_1\right\|\right]\\
            \stackrel{(i)}\leq&\frac{1}{n}\mathbb E\left[\left\|[\td \FS(\wn,\vn)]^{-1}\cdot \sum_{z_i\in U} \nabla_{\w} f(\wn,\vn;z_i)\right\|\right]+\mathbb E[\|\bm{\xi}_1\|]\\
            \stackrel{(ii)}\leq&\frac{1}{n}\cdot\mu^{-1} \mathbb E\left[\left\|\sum_{z_i\in U} \nabla_{\w} f(\wn,\vn;z_i)\right\|\right]+\sqrt{\mathbb E[\|\bm{\xi}_1\|^2]}\\
            \stackrel{(iii)}=&\frac{1}{\mu n} \mathbb E\left[\left\|\sum_{z_i\in U} \nabla_{\w} f(\wn,\vn;z_i)\right\|\right]+\sqrt{d_1}\sigma_1
            \stackrel{(iv)}\leq\frac{mL}{\mu n}+\sqrt{d_1}\sigma_1,
        \end{aligned}
    \end{equation}
    where the inequality ($i$) uses the triangle inequality and the inequality ($ii$) follows by the relation $\td\FS(\wn,\vn) \geq \mu_{\w\w}\geq\mu$, together with the Jensen's inequality to bound $\mathbb E[\|\bm{\xi}_1\|]$. The equality ($iii$) holds because the vector $\bm{\xi}_1\sim\mathcal{N}(0,\sigma_1^2 \mathbf{I}_{d_1})$ and thus we have $\mathbb E[\|\bm{\xi}_1\|^2]=d_1\sigma_1^2$. And the inequality ($iv$) is due to the fact that $f(\w,\v;z)$ is $L$-Lipshitz continuous. Symmetrically, we have
    \begin{equation}\label{eq:eff_vu_vn}
        \mathbb E[\|\tv^u-\vn\|] \leq \frac{mL}{\mu n}+\sqrt{d_2}\sigma_2.
    \end{equation}
    Plugging eq.(\ref{eq:eff_wu_wn}) and eq.(\ref{eq:eff_vu_vn}) into eq.(\ref{eq:eff_PD}) with $d=\max\{d_1,d_2\}$ we get
    \begin{equation}
        \triangle ^w (\tw^u,\tv^u) \leq \frac{2mL^2}{\mu n} + \sqrt{d}(\sigma_1 + \sigma_2)L + \frac{2\sqrt{2}L^2}{\mu n}.
    \end{equation}
    With the noise scale $\sigma_1 $ and $ \sigma_2 $ being equal to $ \frac{2(8 \sqrt{2}L^2 \ell^3 \rho / \mu^6 + 2\sqrt{2}L \ell^2 / \mu^3) m^2}{n^2\epsilon} \sqrt{2\log(2.5/\delta)}$, we can get our generalization guarantee in terms of population weak PD risk:
    \begin{equation}
        \triangle^w (\tw^u,\tv^u) = \mathcal O \left( (L^3 \ell^3 \rho / \mu^6 + L^2 \ell^2/\mu^3)\cdot \frac{m^2 \sqrt{d\log(1/\delta)}}{n^2 \epsilon} + \frac{mL^2}{\mu n}\right).
    \end{equation}
    For the population strong PD risk, using an application of eq.(\ref{eq:unlearning_strPD}) with Lemma \ref{lemma:learning_strPD}, eq.(\ref{eq:eff_wu_wn}), eq.(\ref{eq:eff_vu_vn}) and the noise scales given in Theorem \ref{theorem:certi.eff}, we can get
    \begin{equation}
        \triangle^s (\tw^u,\tv^u) = \mathcal O \left( (L^3 \ell^3 \rho / \mu^6 + L^2 \ell^2/\mu^3)\cdot \frac{m^2 \sqrt{d\log(1/\delta)}}{n^2 \epsilon} + \frac{mL^2}{\mu n} + \frac{L^2\ell}{\mu^2 n}\right).
    \end{equation}
\end{proof}

\subsubsection{Proof of Theorem \ref{theorem:del_cap.eff} (Deletion Capacity)}
\begin{proof}
    By the definition of deletion capacity, in order to ensure the population weak or strong PD risk derived in Lemma \ref{theorem:guarantee_SC.eff} is bounded by $\gamma$, it suffices to let $m ^ { A,\bar A} _ {\epsilon,\delta,\gamma}(d_1,d_2,n) \geq c \cdot \frac{n\sqrt{\epsilon}}{(d\log(1/\delta))^{1/4}}$.
\end{proof}

\subsection{Efficient Online Unlearning Algorithm}
\begin{algorithm}[htbp]
    \caption{Efficient Online Certified Minimax Unlearning $(\bar A_{\tt online})$}
    \label{alg:unlearning.online}
   {\small
    \begin{algorithmic}[1]
    \renewcommand{\algorithmicrequire}{\textbf{Input:}}
    \renewcommand{\algorithmicensure}{\textbf{Output:}}
    \REQUIRE Delete request $z_k \in S$, early delete requests $U$ : $\{z_j\}^m_{j=1}\subseteq S$, output of $A_{sc-sc}(S)$: $(\wn,\vn)$, memory variables $T(S)$: $\{\mathtt D_{\w\w}\FS(\wn,\vn), \mathtt D_{\v\v}\FS(\wn,\vn)\}$, early unlearning variables: $(\tw_U^u, \tv_U^u)$, loss function: $f(\w,\v;z)$, noise parameters: $\sigma_1$, $\sigma_2$.
    
    \STATE Set $\tw_\emptyset = \wn$ and $\tv_\emptyset = \vn$.
    \STATE Compute
        \begin{equation}
            \bw_{U\cup \{k\}} = \tw_U + \frac{1}{n}[\td \FS(\wn,\vn)]^{-1}\nabla_{\w} f(\wn,\vn;z_k),  
        \end{equation}
        \begin{equation}        
            \bv_{U\cup \{k\}} = \tv_U + \frac{1}{n}[\tdv \FS(\wn,\vn)]^{-1}\nabla_{\v} f(\wn,\vn;z_k).  
        \end{equation}
    \STATE $\tw^u_{U \cup \{k\}}=\bw_{U\cup \{k\}}+\bm{\xi}_1$, where $\bm{\xi}_1\sim \mathcal N(0,\sigma_1\mathbf{I}_{d_1})$ and $\tv^u_{U \cup \{k\}}=\bv_{U\cup \{k\}}+\bm{\xi}_2$, where $\bm{\xi}_2\sim \mathcal N(0,\sigma_2\mathbf{I}_{d_2})$ .
    \ENSURE $(\tw^u_{U \cup \{k\}},\tv^u_{U \cup \{k\}})$.
     \end{algorithmic}  
     }
\end{algorithm}
To support successive unlearning requests, similar to the STL case \citep{suriyakumar2022algorithms}, we further provide an efficient and online minimax unlearning algorithm (denoted by $\bar A_{\tt online}$). The pseudocode of $\bar A_{\tt online}$ is given in Algorithm \ref{alg:unlearning.online}. Its certified minimax unlearning guarantee, generalization, and deletion capacity can be identically yielded as Algorithm \ref{alg:unlearning.extension.online}, which are omitted here.
\section{Detailed Algorithm Descriptions and Missing Proofs in Section 5}\label{appendix:C}
\subsection{Minimax Unlearning Algorithm for Smooth Convex-Concave Loss Function}
\label{appendix:C.1}
In this section, we provide minimax learning and minimax unlearning algorithms for smooth convex-concave loss functions based on the counterpart algorithms for the SC-SC setting. Given the convex-concave loss function $f(\w,\v;z)$, we define the regularized loss function as $\widetilde f(\w,\v;z) = f(\w,\v;z) + \frac{\lambda}{2}\|\w\|^2 - \frac{\lambda}{2}\|\v\|^2$. Suppose the function $f$ satisfies Assumption \ref{ass:1}, then the function $\widetilde f$ is $\lambda$-strongly convex in $\w$, $\lambda$-strongly concave in $\v$, $(2L+\lambda\|\w\|+\lambda\|\v\|)$-Lipschitz, $\sqrt{2}(2\ell+\lambda)$-gradient Lipschitz and $\rho$-Hessian Lipschitz. Thus, we can apply the minimax learning in Algorithm \ref{alg:learning} and unlearning in Algorithm \ref{alg:unlearning} to the regularized loss function with a properly chosen $\lambda$. We denote our learning algorithm by $A_{c-c}$ and unlearning algorithm by $\bar A_{c-c}$. The pseudocode is provided in Algorithm \ref{alg:learning_C} and Algorithm \ref{alg:unlearning_C}, respectively. Additionally, we denote the regularized population loss as $\widetilde F(\w,\v):=\mathbb E_{z\sim\mathcal D}[\widetilde f(\w,\v;z)]$ and regularized empirical loss as $\WFS(\w,\v):= \frac{1}{n}\sum_{i=1}^n \widetilde f(\w,\v;z_i)$.

\begin{algorithm}[htbp]
    \caption{Mimimax Learning Algorithm $(A_{c-c})$}
    \label{alg:learning_C}
    {\small
    \begin{algorithmic}[1]
    \renewcommand{\algorithmicrequire}{\textbf{Input:}}
    \renewcommand{\algorithmicensure}{\textbf{Output:}}
    \REQUIRE Dataset $S$ : $\{z_i\}^n_{i=1}\sim\mathcal D^n$, loss function: $f(\w,\v;z)$, regularization parameter: $\lambda$.
    \STATE Define 
    \begin{equation}
        \widetilde f(\w,\v;z) = f(\w,\v;z) + \frac{\lambda}{2}\|\w\|^2 - \frac{\lambda}{2}\|\v\|^2.        
    \end{equation}
    \STATE Run the algorithm $A_{sc-sc}$ on the dataset $S$ with loss function $\widetilde f$.
    \ENSURE $(\wn, \vn, \mathtt D_{\w\w}\WFS(\wn,\vn), \mathtt D_{\v\v}\WFS(\wn,\vn)) \leftarrow A_{sc-sc}(S, \widetilde f)$.

     \end{algorithmic}  
    }
\end{algorithm}
\begin{algorithm}[htbp]
    \caption{Certified Minimax Unlearning for Convex-Concave Loss $(\bar A_{c-c})$}
    \label{alg:unlearning_C}
    {\small
    \begin{algorithmic}[1]
    \renewcommand{\algorithmicrequire}{\textbf{Input:}}
    \renewcommand{\algorithmicensure}{\textbf{Output:}}
    \REQUIRE Delete requests $U$ : $\{z_j\}^m_{j=1}\subseteq S$, output of $A_{c-c}(S)$: $(\wn,\vn)$, loss function: $f(\w,\v;z)$, memory variables $T(S)$: $\{\mathtt D_{\w\w}\WFS(\wn,\vn), \mathtt D_{\v\v}\WFS(\wn,\vn)\}$, regularization parameter: $\lambda$, noise parameters: $\sigma_1$, $\sigma_2$.
    \STATE Define 
    \begin{equation}
        \widetilde f(\w,\v;z) = f(\w,\v;z) + \frac{\lambda}{2}\|\w\|^2 - \frac{\lambda}{2}\|\v\|^2.        
    \end{equation}
    \STATE Run the algorithm $\bar A_{sc-sc}$ with delete requests $U$, learning variables $(\wn,\vn)$, memory variables $T(S)$, loss function $\widetilde f$ and noise parameters $\sigma_1$ and $\sigma_2$.
    \ENSURE $(\wu,\vu) \leftarrow \bar A_{sc-sc}(U, (\wn,\vn), T(S), \widetilde f, \sigma_1, \sigma_2)$.
     \end{algorithmic}  
     }
\end{algorithm}

\subsection{Supporting Lemma}
\begin{lemma}\label{lemma:empirical_bound}
    Suppose the function $f(\w,\v;z)$ is $L$-Lipschitz continuous. Define the function $\widetilde f(\w,\v;z)$ as
    \begin{equation}
        \widetilde f(\w,\v;z) = f(\w,\v;z) + \frac{\lambda}{2}\|\w\|^2 - \frac{\lambda}{2}\|\v\|^2. 
    \end{equation}
    Given a dataset $S=\{z_i\}^n_{i=1}$ and denote $(\wn,\vn) := \arg\min_{\w\in\mathcal W}\max_{\v\in\mathcal V}\{\WFS(\w,\v):=\frac{1}{n}\sum_{i=1}^n \widetilde f(\w,\v;z_i)\}$. Then, the variables $(\wn,\vn)$ satisfy $\|\wn\|\leq L/\lambda$ and $\|\vn\| \leq L/\lambda$.
\end{lemma}
\begin{proof}
    Due to the optimality of $\wn$, we have
    \begin{equation}
        \nabla_{\w}\WFS(\wn,\vn;z) = \frac{1}{n}\sum_{z_i\in S}\nabla_{\w}\widetilde f(\wn,\vn;z_i) = 0.
    \end{equation}
    Plugging in the definition of the function $\widetilde f$ in the above, we get that
    \begin{equation}
        \frac{1}{n}\sum_{z_i\in S}\nabla_{\w}f(\wn,\vn;z_i) + \lambda \wn= 0.
    \end{equation}
    Then, using the triangle inequality, we have
    \begin{equation}
        \|\lambda \wn\| = \| \frac{1}{n}\sum_{z_i\in S}\nabla_{\w}f(\wn,\vn;z_i) \| \leq \frac{1}{n}\sum_{z_i\in S} \|\nabla_{\w}f(\wn,\vn;z_i) \| \leq L,
    \end{equation}
    where the last inequality holds because the function $f$ is $L$-Lipschitz continuous. Thus we have $\|\wn\|\leq L/\lambda$. Similarly, we can get $\|\vn\| \leq L/\lambda$.
\end{proof}
Lemma \ref{lemma:empirical_bound} implies that the empirical optimizer $(\wn,\vn)$ returned by Algorithm \ref{alg:unlearning_C} satisfies $\|\wn\|\leq L/\lambda$ and $\|\vn\| \leq L/\lambda$. Thus our domain of interest are $\mathcal W:=\{\w|\|\w\|\leq L/\lambda\}$ and $\mathcal V:=\{\v|\|\v\|\leq L/\lambda\}$. Over the set $\mathcal W\times \mathcal V$, the function $\widetilde f(\w,\v;z)$ is $4L$-Lipschitz continuous. Also, with $\lambda < \ell$, $\widetilde f(\w,\v;z)$ has $3\sqrt{2} \ell$-Lipschitz gradients.

\subsection{Proof of Theorem \ref{theorem:guarantee_C} (Certified Minimax Unlearning for Convex-Concave Loss Funcion)}
Denote $\widetilde L = 4L$ and $\widetilde \ell = 3\sqrt{2} \ell$, then the function $\widetilde f$ is $\widetilde L$-Lipschitz continuous and has $\widetilde \ell$-Lipschitz gradients. Let $(\wnu,\vnu)$ be the optimal solution of the loss function $\WFU(\w,\v)$ on the remaining dataset, i.e.,
\begin{equation}
    \label{eq:remain_optimal_C}
        (\wnu,\vnu) := \arg\min_{\w\in\mathcal W}\max_{\v\in\mathcal V}\{\WFU(\w,\v):=\frac{1}{n-m}\sum_{z_i\in S\setminus U}\widetilde f(\w,\v;z_i)\}.
\end{equation}
Additionally, we have $\| \td \WFS(\wn,\vn) \| \geq \lambda$ and $\| \tdv \WFS(\wn,\vn) \| \geq \lambda$.
\begin{lemma}\label{lemma:unlearning_PD_C}
    Under the settings of Theorem \ref{theorem:guarantee_C}, for any $\lambda>0$, the population weak and strong PD risk for the minimax learning variables $(\wn,\vn)$ returned by Algorithm \ref{alg:learning_C} are
    \begin{equation}
    \left \{
    \begin{aligned}
        & \max_{\v\in \mathcal V} \mathbb E[F(\wn,\v)] - \min_{\w\in\mathcal W} \mathbb E[F(\w,\vn)] \leq \frac{32\sqrt{2}L^2}{\lambda n} +\frac{\lambda}{2}(B_{\w}^2+ B_{\v}^2), \\    
        & \mathbb E[\max_{\v\in \mathcal V} F(\wn, \v) - \min_{\w\in \mathcal W} F(\w, \vn)] \leq \frac{384\sqrt{2} L^2 \ell}{\lambda^2 n} + \frac{\lambda}{2}(B_{\w}^2+ B_{\v}^2).
    \end{aligned}
    \right .
    \end{equation}
\end{lemma}
\begin{proof}
    For the function $\widetilde F(\w, \v)$, an application of Lemma \ref{lemma:learning_PD} gives that
    \begin{equation}
        \max_{\v\in \mathcal V} \mathbb E[\widetilde F(\wn,\v)] - \min_{\w\in\mathcal W} \mathbb E[\widetilde F(\w,\vn)]\leq \frac{32\sqrt{2}L^2}{\lambda n}.
    \end{equation}
    By the assumption of bounded parameter spaces $\mathcal W$ and $\mathcal V$ so that $\max_{\w\in\mathcal W}\Vert\w\Vert\leq B_{\w}$ and $\max_{\v\in\mathcal V}\Vert\v\Vert\leq B_{\v}$, we have the following derivations for the population weak PD risk,
    \begin{equation}\label{eq:unlearning_wPD_C}
        \begin{split}
            &\max_{\v\in \mathcal V} \mathbb E[F(\wn,\v)] - \min_{\w\in\mathcal W} \mathbb E[F(\w,\vn)] \\
            =& \max_{\v\in \mathcal V} \mathbb E\bigg[\widetilde F(\wn,\v) -\frac{\lambda}{2}\|\wn\|^2+\frac{\lambda}{2}\|\v\|^2\bigg] - \min_{\w\in\mathcal W} \mathbb E\bigg[\widetilde F(\w,\vn)-\frac{\lambda}{2}\|\w\|^2+\frac{\lambda}{2}\|\vn\|^2\bigg] \\
            \leq & \max_{\v\in \mathcal V} \mathbb E\bigg[\widetilde F(\wn,\v) +\frac{\lambda}{2}\|\v\|^2\bigg] - \min_{\w\in\mathcal W} \mathbb E\bigg[\widetilde F(\w,\vn)-\frac{\lambda}{2}\|\w\|^2\bigg] \\
            \leq & \max_{\v\in \mathcal V} \mathbb E\big[\widetilde F(\wn,\v)\big] - \min_{\w\in\mathcal W} \mathbb E\big[\widetilde F(\w,\vn)\big] + \max_{\v\in \mathcal V} \mathbb E\bigg[\frac{\lambda}{2}\|\v\|^2\bigg] + \max_{\w\in\mathcal W}\mathbb E\bigg[\frac{\lambda}{2}\|\w\|^2\bigg] \\
            \leq & \frac{32\sqrt{2}L^2}{\lambda n} +\frac{\lambda}{2}(B_{\w}^2+ B_{\v}^2).
        \end{split}
    \end{equation}
    Similarly, an application of Lemma \ref{lemma:learning_strPD} gives that
    \begin{equation}
        \mathbb E[\max_{\v\in \mathcal V} \widetilde F(\wn, \v) - \min_{\w\in \mathcal W} \widetilde F(\w, \vn)] \leq \frac{128 \sqrt{2} L^2 (2\ell + \lambda)}{\lambda^2 n}.
    \end{equation}
    And we can get the population strong PD risk with the following conversions,
    \begin{equation}\label{eq:unlearning_strPD_C}
        \begin{aligned}
            & \mathbb E[\max_{\v\in \mathcal V} F(\wn, \v) - \min_{\w\in \mathcal W} F(\w, \vn)] \\
            = & \mathbb E\left [\max_{\v\in \mathcal V} \left( \WF(\wn, \v) -\frac{\lambda}{2}\|\wn\|^2+\frac{\lambda}{2}\|\v\|^2\right) - \min_{\w\in \mathcal W} \left( \WF(\w, \vn) -\frac{\lambda}{2}\|\w\|^2+\frac{\lambda}{2}\|\vn\|^2 \right) \right] \\
            \leq & \mathbb E\left [\max_{\v\in \mathcal  V} \left( \WF(\wn, \v) + \frac{\lambda}{2} \|\v\|^2 \right) - \min_{\w\in \mathcal W} \left( \WF(\w, \vn) - \frac{\lambda}{2} \|\w\|^2 \right) \right] \\
            = & \mathbb E\left [\max_{\v\in \mathcal V} \left( \WF(\wn, \v) + \frac{\lambda}{2} \|\v\|^2 \right) \right] - \mathbb E\left [\min_{\w\in \mathcal W} \left( \WF(\w, \vn) - \frac{\lambda}{2} \|\w\|^2 \right) \right] \\
            \leq & \mathbb E[\max_{\v\in \mathcal  V} \WF(\wn, \v) + \max_{\v\in \mathcal  V} \frac{\lambda}{2} \|\v\|^2] + \mathbb E[\max_{\w\in \mathcal W} (-\WF)(\w, \vn) + \max_{\w\in \mathcal W} \frac{\lambda}{2} \|\w\|^2]\\
            = & \mathbb E [\max_{\v\in \mathcal  V} \WF(\wn, \v) -\min_{\w\in \mathcal W} \WF(\w, \vn)] + \mathbb E[\max_{\v\in \mathcal  V} \frac{\lambda}{2} \|\v\|^2] + \mathbb E[\max_{\w\in \mathcal W} \frac{\lambda}{2} \|\w\|^2] \\
            \leq & \frac{128 \sqrt{2} L^2 (2\ell + \lambda)}{\lambda^2 n} + \frac{\lambda}{2}(B_{\w}^2+ B_{\v}^2) \leq \frac{384\sqrt{2} L^2 \ell}{\lambda^2 n} + \frac{\lambda}{2}(B_{\w}^2+ B_{\v}^2).
        \end{aligned}
    \end{equation}
\end{proof}

\begin{lemma}[\textbf{Closeness Upper Bound}]
\label{lemma:sensitivity_C}
    Under the settings of Theorem \ref{theorem:guarantee_C}, we have the closeness bound between the intermediate variables $(\widehat{\w},\widehat{\v})$ in Algorithm \ref{alg:unlearning_C} and $(\wnu,\vnu)$ in eq.(\ref{eq:remain_optimal_C}):
    \begin{equation}
       \{ \|\wnu-\hw\|, \|\vnu-\hv\|\} \leq \frac{(8 \sqrt{2}\widetilde L^2 \widetilde \ell^3 \rho / \lambda^5 + 8 \widetilde L \widetilde \ell^2 / \lambda^2) m^2}{n ( \lambda n - (\widetilde \ell + \widetilde \ell^2/\lambda)m )}.
    \end{equation}
\end{lemma}
\begin{proof}
    Since we now run the algorithms $A_{sc-sc}$ and $\bar A_{sc-sc}$ with the regularized loss function $\widetilde f$, the proof is identical to that of Lemma \ref{lem:sensitivity}.
\end{proof}

Equipped with the supporting lemmas above, the proof of Theorem \ref{theorem:guarantee_C} can be separated into the proofs of the following three lemmas.
\begin{lemma}[\textbf{Minimax Unlearning Certification}]
\label{lemma:certi_C}
    Under the settings of Theorem \ref{theorem:guarantee_C}, our minimax learning algorithm $A_{c-c}$ and unlearning algorithm $\bar A_{c-c}$ is $(\epsilon,\delta)$-certified minimax unlearning if we choose
    \begin{equation}\label{eq:noise_C}
        \sigma_1 \text{~and~} \sigma_2 = \frac{2(8 \sqrt{2}\widetilde L^2 \widetilde \ell^3 \rho / \lambda^5 + 8 \widetilde L \widetilde \ell^2 / \lambda^2) m^2}{n ( \lambda n - (\widetilde \ell + \widetilde\ell^2/\lambda)m ) \epsilon} \sqrt{2\log(2.5/\delta)}.
    \end{equation}
\end{lemma}
\begin{proof}
    With the closeness upper bound in Lemma \ref{lemma:sensitivity_C} and the given noise scales in eq.(\ref{eq:noise_C}), the proof is identical to that of Theorem \ref{theorem:certi}.
\end{proof}
\begin{lemma}[\textbf{Population Primal-Dual Risk}]
\label{lemma:risk_C}
    Under the settings of Theorem \ref{theorem:guarantee_C}, the population weak and strong PD risk for $(\wu,\vu)$ returned by Algorithm \ref{alg:unlearning_C} are
    \begin{equation}
    \left \{
        \begin{aligned}
            & \triangle^w(\wu,\vu) \leq \mathcal O \bigg( (L^3 \ell^3 \rho / \lambda^6 + L^2 \ell^2/\lambda^3)\cdot \frac{m^2 \sqrt{d\log(1/\delta)}}{n^2 \epsilon} + \frac{mL^2}{\lambda n} + \lambda(B_{\w}^2 + B_{\v}^2) \bigg), \\
            & \triangle^s(\wu,\vu) \leq \mathcal O \bigg( (L^3 \ell^3 \rho / \lambda^6 + L^2 \ell^2/\lambda^3)\cdot \frac{m^2 \sqrt{d\log(1/\delta)}}{n^2 \epsilon} + \frac{mL^2}{\lambda n} + \frac{L^2 \ell}{\lambda^2 n} + \lambda(B_{\w}^2 + B_{\v}^2) \bigg).
        \end{aligned}
    \right .
    \end{equation}
\end{lemma}
\begin{proof}
    For the population weak PD risk, an application of eq.(\ref{eq:unlearning_PD}) together with Lemma \ref{lemma:unlearning_PD_C} gives that
    \begin{equation}\label{eq:PD_C}
        \triangle^w(\wu,\vu) \leq \mathbb E[L \|\wu-\wn\|] + \mathbb E[L \|\vu-\vn\|] + \frac{32\sqrt{2}L^2}{\lambda n} +\frac{\lambda}{2}(B_{\w}^2+ B_{\v}^2).
    \end{equation}
    According to Algorithm \ref{alg:unlearning_C}, we have the unlearning update step
    \begin{equation}
    \label{eq:wu_C}
        \wu = \wn+\frac{1}{n-m}[\td \WFU(\wn,\vn)]^{-1}\sum_{z_i\in U}\nabla_{\w} \widetilde f(\wn,\vn;z_i)+\bm{\xi}_1,
    \end{equation}
    where $\WFU(\wn,\vn):=\frac{1}{n-m}\sum_{z_i\in S\setminus U}\widetilde f(\w,\v;z_i)$. From the relation above, we further get
    \begin{equation}\label{eq:wu_wn_C}
        \begin{aligned}
            & \mathbb E[\|\wu-\wn\|]\\
            =&\mathbb E\left[\left\|\frac{1}{n-m}[\td \WFU(\wn,\vn)]^{-1}\cdot \sum_{z_i\in U} \nabla_{\w} \widetilde f(\wn,\vn;z_i) + \bm{\xi}_1\right\|\right]\\
            \stackrel{(i)}\leq&\frac{1}{n-m}\mathbb E\left[\left\|[\td \WFU(\wn,\vn)]^{-1}\cdot \sum_{z_i\in U} \nabla_{\w} \widetilde f(\wn,\vn;z_i)\right\|\right]+\mathbb E[\|\bm{\xi}_1\|]\\
            \stackrel{(ii)}\leq&\frac{1}{n-m}\cdot\frac{n-m}{(\lambda n - \widetilde\ell (1+\widetilde\ell/\lambda)m )} \mathbb E\left[\left\|\sum_{z_i\in U} \nabla_{\w} \widetilde f(\wn,\vn;z_i)\right\|\right]+\sqrt{\mathbb E[\|\bm{\xi}_1\|^2]}\\
            \stackrel{(iii)}=&\frac{1}{(\lambda n - \widetilde\ell (1+\widetilde\ell/\lambda)m )} \mathbb E\left[\left\|\sum_{z_i\in U} \nabla_{\w} \widetilde f(\wn,\vn;z_i)\right\|\right]+\sqrt{d_1}\sigma_1\\
            \stackrel{(iv)}\leq & \frac{1}{(\lambda n - \widetilde\ell (1+\widetilde\ell/\lambda)m )} \mathbb E \left[ \sum_{z_i\in U} \big(\|\nabla_{\w} f(\wn,\vn;z_i)\| + \lambda\|\wn\| + \lambda\|\vn\| \big)\right]+\sqrt{d_1}\sigma_1\\
            \stackrel{(vi)}\leq & \frac{3mL}{(\lambda n - \widetilde\ell (1+\widetilde\ell/\lambda)m )}+\sqrt{d_1}\sigma_1,
        \end{aligned}
    \end{equation}
    where the inequality ($i$) uses the triangle inequality and the inequality ($ii$) follows from an application of eq.(\ref{eq:maineq_right}), together with the Jensen's inequality to bound $\mathbb E[\|\bm{\xi}_1\|]$. The equality ($iii$) holds because the vector $\bm{\xi}_1\sim\mathcal{N}(0,\sigma_1^2 \mathbf{I}_{d_1})$ and thus we have $\mathbb E[\|\bm{\xi}_1\|^2]=d_1\sigma_1^2$. The inequality ($iv$) uses the definition of the function $\widetilde f$ and the triangle inequality. The inequality ($vi$) is due to the fact that $f(\w,\v;z)$ is $L$-Lipshitz continuous and Lemma \ref{lemma:empirical_bound}. Symmetrically, we have
    \begin{equation}\label{eq:vu_vn_C}
        \mathbb E[\|\vu-\vn\|] \leq \frac{3mL}{(\lambda n - \widetilde\ell (1+\widetilde\ell/\lambda)m )}+\sqrt{d_2}\sigma_2.
    \end{equation}
    Plugging eq.(\ref{eq:wu_wn_C}) and eq.(\ref{eq:vu_vn_C}) into eq.(\ref{eq:PD_C}) with noise scales given in Lemma \ref{lemma:certi_C}, we can get our generalization guarantee in terms of population weak PD risk:
    \begin{equation}
        \triangle^w(\wu,\vu) \leq \mathcal O \bigg( (L^3 \ell^3 \rho / \lambda^6 + L^2 \ell^2/\lambda^3)\cdot \frac{m^2 \sqrt{d\log(1/\delta)}}{n^2 \epsilon} + \frac{mL^2}{\lambda n} + \lambda(B_{\w}^2 + B_{\v}^2) \bigg).
    \end{equation}
    Similarly, using an application of eq.(\ref{eq:unlearning_strPD}) together with Lemma \ref{lemma:unlearning_PD_C}, Lemma \ref{lemma:certi_C}, eq.(\ref{eq:wu_wn_C}) and eq.(\ref{eq:vu_vn_C}), we can get the following population strong PD risk: 
    \begin{equation}
        \triangle^s(\wu,\vu) \leq \mathcal O \bigg( (L^3 \ell^3 \rho / \lambda^6 + L^2 \ell^2/\lambda^3)\cdot \frac{m^2 \sqrt{d\log(1/\delta)}}{n^2 \epsilon} + \frac{mL^2}{\lambda n} + \frac{L^2 \ell}{\lambda^2 n} + \lambda(B_{\w}^2 + B_{\v}^2) \bigg).
    \end{equation}
\end{proof}
\begin{lemma}[\textbf{Deletion Capacity}]
    Under the settings of Theorem \ref{theorem:guarantee_C}, the deletion capacity of Algorithm \ref{alg:unlearning_C} is
    \begin{equation}
        m ^ { A,\bar A} _ {\epsilon,\delta,\gamma}(d_1,d_2,n) \geq c \cdot \frac{n\sqrt{\epsilon}}{(d\log(1/\delta))^{1/4}},
    \end{equation}
    where the constant $c$ depends on $L, l,\rho, B_{\w}$ and $B_{\v}$. 
\end{lemma}
\begin{proof}
    By the definition of deletion capacity, in order to ensure the population PD risk derived in Lemma \ref{lemma:risk_C} is bounded by $\gamma$, it suffices to let $m ^ { A,\bar A} _ {\epsilon,\delta,\gamma}(d_1,d_2,n) \geq c \cdot \frac{n\sqrt{\epsilon}}{(d\log(1/\delta))^{1/4}}$.
\end{proof}
\subsection{Minimax Unlearning Algorithm for Smooth Convex-Strongly-Concave Loss Function}
In this section, we briefly discuss the extension to the smooth C-SC setting. The SC-C setting is symmetric and thus omitted here. 

Given the loss function $f(\w,\v;z)$ that satisfies Assumption \ref{ass:1} with $\mu_{\v}$-strong concavity in $\v$, we define the regularized function as $\widetilde f(\w,\v;z) = f(\w,\v;z) + \frac{\lambda}{2}\|\w\|^2$. Our minimax learning and minimax unlearning algorithms for C-SC loss function $f$ denoted by $A_{c-sc}$ and $\bar A_{c-sc}$ are given in Algorithm \ref{alg:learning_CSC} and Algorithm \ref{alg:unlearning_CSC} respectively. Additionally, we denote the regularized population loss by $\widetilde F(\w,\v):=\mathbb E_{z\sim\mathcal D}[\widetilde f(\w,\v;z)]$ and regularized empirical loss by $\WFS(\w,\v):= \frac{1}{n}\sum_{i=1}^n \widetilde f(\w,\v;z_i)$.
\begin{algorithm}[htbp]
    \caption{Mimimax Learning Algorithm $(A_{c-sc})$}
    \label{alg:learning_CSC}
    {\small
    \begin{algorithmic}[1]
    \renewcommand{\algorithmicrequire}{\textbf{Input:}}
    \renewcommand{\algorithmicensure}{\textbf{Output:}}
    \REQUIRE Dataset $S$ : $\{z_i\}^n_{i=1}\sim\mathcal D^n$, loss function: $f(\w,\v;z)$, regularization parameter: $\lambda$.
    \STATE Define 
    \begin{equation}
        \widetilde f(\w,\v;z) = f(\w,\v;z) + \frac{\lambda}{2}\|\w\|^2.        
    \end{equation}
    \STATE Run the algorithm $A_{sc-sc}$ on the dataset $S$ with loss function $\widetilde f$.
    \ENSURE $(\wn, \vn, \mathtt D_{\w\w}\WFS(\wn,\vn), \mathtt D_{\v\v}\WFS(\wn,\vn)) \leftarrow A_{sc-sc}(S, \widetilde f)$.

     \end{algorithmic}  
    }
\end{algorithm}
\begin{algorithm}[htbp]
    \caption{Certified Minimax Unlearning for Convex-Strongly-Concave Loss $(\bar A_{c-sc})$}
    \label{alg:unlearning_CSC}
    {\small
    \begin{algorithmic}[1]
    \renewcommand{\algorithmicrequire}{\textbf{Input:}}
    \renewcommand{\algorithmicensure}{\textbf{Output:}}
    \REQUIRE Delete requests $U$ : $\{z_j\}^m_{j=1}\subseteq S$, output of $A_{c-sc}(S)$: $(\wn,\vn)$, memory variables $T(S)$: $\{\mathtt D_{\w\w}\WFS(\wn,\vn), \mathtt D_{\v\v}\WFS(\wn,\vn)\}$, loss function: $f(\w,\v;z)$, regularization parameter: $\lambda$, noise parameters: $\sigma_1$, $\sigma_2$.
    \STATE Define 
    \begin{equation}
        \widetilde f(\w,\v;z) = f(\w,\v;z) + \frac{\lambda}{2}\|\w\|^2.
    \end{equation}
    \STATE Run the algorithm $\bar A_{sc-sc}$ with delete requests $U$, learning variables $(\wn,\vn)$, memory variables $T(S)$, loss function $\widetilde f$ and noise parameters $\sigma_1$ and $\sigma_2$.
    \ENSURE $(\wu,\vu) \leftarrow \bar A_{sc-sc}(U, (\wn,\vn), T(S), \widetilde f, \sigma_1, \sigma_2)$.
     \end{algorithmic}  
     }
\end{algorithm}

Note that the function $\widetilde f(\w,\v;z)$ is $\lambda$-strongly convex in $\w$, $\mu_{\v}$-strongly concave in $\v$, $(\widetilde
 L:=2L+\lambda\|\w\|)$-Lipschitz, $(\widetilde\ell:=\sqrt{2}(2\ell+\lambda))$-gradient Lipschitz and $\rho$-Hessian Lipschitz. We also have $\| \td \WFS(\wn,\vn) \| \geq \lambda$. Let $(\wnu,\vnu)$ be the optimal solution of the loss function $\WFU(\w,\v)$ on the remaining dataset, i.e.,
\begin{equation}
    \label{eq:remain_optimal_CSC}
        (\wnu,\vnu) := \arg\min_{\w\in\mathcal W}\max_{\v\in\mathcal V}\{\WFU(\w,\v):=\frac{1}{n-m}\sum_{z_i\in S\setminus U}\widetilde f(\w,\v;z_i)\}.
\end{equation}
An application of Lemma \ref{lemma:empirical_bound} implies that the empirical optimizer $(\wn,\vn)$ returned by Algorithm \ref{alg:unlearning_CSC} satisfies $\|\wn\|\leq L/\lambda$. Thus our domain of interest are $\mathcal W:=\{\w|\|\w\|\leq L/\lambda\}$. Over the set $\mathcal W$, the function $\widetilde f(\w,\v;z)$ is $3L$-Lipschitz continuous. Suppose the strongly-convex regularization parameter $\lambda$ satisfies $\lambda<\ell$, then $\widetilde f(\w,\v;z)$ has $3\sqrt{2} \ell$-Lipschitz gradients.

The corresponding theoretical results are given below.
\begin{lemma}[\textbf{Closeness Upper Bound}]
\label{lemma:sensitivity_CSC}
    Let Assumption \ref{ass:1} hold. Assume the function $f(\w,\v;z)$ is $\mu_{\v}$-strongly concave in $\v$ and $\|\tdv \WFS(\wn,\vn)\|\geq \mu_{\v\v}$. Let $\mu = \min\{\mu_{\v},\mu_{\v\v}\}$. Then, we have the closeness bound between the intermediate variables $(\widehat{\w},\widehat{\v})$ in Algorithm \ref{alg:unlearning_CSC} and $(\wnu,\vnu)$ in eq.(\ref{eq:remain_optimal_CSC}):
\begin{equation}
    \left\{
        \begin{array}{lr}
            \|\wnu-\hw\| \leq \big( \frac{8\sqrt{2}\widetilde L^2 \widetilde\ell^3 \rho}{\lambda^2 \mu^3} + \frac{8\widetilde L\widetilde\ell^2}{\lambda \mu} \big)\cdot\frac{m^2}{n(\lambda n- (\widetilde \ell + \widetilde \ell^2/\mu)m)},\\
            \|\vnu-\hv\| \leq \big( \frac{8\sqrt{2}\widetilde L^2 \widetilde\ell^3 \rho}{\lambda^3 \mu^2} + \frac{8\widetilde L\widetilde\ell^2}{\lambda \mu} \big)\cdot\frac{m^2}{n(\mu n- (\widetilde \ell + \widetilde \ell^2/\lambda)m)}.
        \end{array}
    \right.
\end{equation}  
\end{lemma}
\begin{proof}
    Since we now run the algorithms $A_{sc-sc}$ and $\bar A_{sc-sc}$ with the regularized loss function $\widetilde f$, the proof is identical to that of Lemma \ref{lem:sensitivity}.
\end{proof}

\begin{lemma}[\textbf{Minimax Unlearning Certification}]
\label{lemma:certi_CSC}
    Under the settings of Lemma \ref{lemma:sensitivity_CSC}, our minimax learning algorithm $A_{c-sc}$ and unlearning algorithm $\bar A_{c-sc}$ is $(\epsilon,\delta)$-certified minimax unlearning if we choose
    \begin{equation}\label{eq:noise_CSC}
        \left\{
            \begin{array}{lr}
                 \sigma_1 = \big( \frac{8\sqrt{2}\widetilde L^2 \widetilde\ell^3 \rho}{\lambda^2 \mu^3} + \frac{8\widetilde L\widetilde\ell^2}{\lambda \mu} \big)\cdot\frac{2m^2 \sqrt{2\log(2.5/\delta)}}{n(\lambda n- (\widetilde \ell + \widetilde \ell^2/\mu)m)\epsilon},\\
                 \sigma_2 = \big( \frac{8\sqrt{2}\widetilde L^2 \widetilde\ell^3 \rho}{\lambda^3 \mu^2} + \frac{8\widetilde L\widetilde\ell^2}{\lambda \mu} \big)\cdot\frac{2m^2 \sqrt{2\log(2.5/\delta)}}{n(\mu n- (\widetilde \ell + \widetilde \ell^2/\lambda)m)\epsilon}.
            \end{array}
        \right .
    \end{equation}
\end{lemma}
\begin{proof}
    With the closeness upper bound in Lemma \ref{lemma:sensitivity_CSC} and the given noise scales in eq.(\ref{eq:noise_CSC}), the proof is identical to that of Theorem \ref{theorem:certi}.
\end{proof}

\begin{lemma}[\textbf{Population Weak PD Risk}]
\label{lemma:risk_CSC}
    Under the same settings of Lemma \ref{lemma:sensitivity_CSC}, suppose the parameter space $\mathcal W$ is bounded so that $\max_{\w\in\mathcal W}\|\w\|\leq B_{\w}$, the population weak PD risk for the certified minimax unlearning variables $(\wu,\vu)$ returned by Algorithm \ref{alg:unlearning_CSC} is
    \begin{equation}
        \triangle^w(\wu,\vu) \leq \mathcal O \bigg( \big(\frac{L^3 \ell^3 \rho}{\lambda^3 \mu^3} + \frac{L^2 \ell^2}{\lambda^2 \mu} + \frac{L^2 \ell^2}{\lambda \mu^2}\big)\cdot\frac{m^2 \sqrt{d\log(1/\delta)}}{n^2 \epsilon} + \frac{mL^2}{\lambda n} +\frac{mL^2}{\mu n} + \lambda B_{\w}^2 \bigg),
    \end{equation}
    where $d=\max\{d_1,d_2\}$. In particular, by setting the regularization parameter $\lambda$ as:
    \begin{equation}
    \begin{split}
        \lambda = \max \bigg\{ & \frac{L}{B_{\w}}\sqrt{\frac{m}{n}}, \frac{L\ell m}{B_{\w}\mu n} \big(\frac{\sqrt{d\log(1/\delta)}}{\epsilon}\big)^{1/2}, \big(\frac{L^2\ell^2m^2\sqrt{d\log(1/\delta)}}{B_{\w}^2\mu n^2\epsilon}\big)^{1/3},
        \big( \frac{L^3\ell^3\rho m^2\sqrt{d\log(1/\delta)}}{B_{\w}^2\mu^3n^2\epsilon} \big)^{1/4} \bigg\},        
    \end{split}
    \end{equation}
    we have the following population weak PD risk:
    \begin{equation}
    \begin{split}
        \triangle^w(\wu,\vu) \leq \mathcal O \bigg( & c_1 \sqrt{\frac{m}{n}} + c_2\frac{m}{n} + c_3 \big(\frac{\sqrt{d\log(1/\delta)}}{\epsilon}\big)^{1/2}\frac{m}{n} \\
        & + c_4 \big(\frac{\sqrt{d\log(1/\delta)}}{\epsilon}\big)^{1/3} \big(\frac{m}{n}\big)^{2/3} + c_5 \big(\frac{\sqrt{d\log(1/\delta)}}{\epsilon}\big)^{1/4} \sqrt{\frac{m}{n}} \bigg),
    \end{split}
    \end{equation}
    where $c_1, c_2, c_3, c_4$ and $c_5$ are constants that depend only on $L, l, \rho, \mu$ and $B_{\w}$.
\end{lemma}
\begin{proof}
    An application of \citep[Theorem 1]{zhang2021generalization} gives that
    \begin{equation}
        \max_{\v\in \mathcal V} \mathbb E[\widetilde F(\wn,\v)] - \min_{\w\in\mathcal W} \mathbb E[\widetilde F(\w,\vn)]\leq \frac{18\sqrt{2}L^2}{n}\bigg(\frac{1}{\lambda}+\frac{1}{\mu}\bigg).
    \end{equation}
    Using the relation above with an application of eq.(\ref{eq:unlearning_PD}) and eq.(\ref{eq:unlearning_wPD_C}), we have
    \begin{equation}\label{eq:PD.CSC}
        \triangle^w(\wu,\vu) \leq \mathbb E[L \|\wu-\wn\|] + \mathbb E[L \|\vu-\vn\|] + \frac{18\sqrt{2}L^2}{ n}\bigg(\frac{1}{\lambda}+\frac{1}{\mu}\bigg) + \frac{\lambda B_{\w}^2}{2}.
    \end{equation}
    By an application of eq.(\ref{eq:wu_wn_C}), we further get
    \begin{equation}\label{eq:wu.CSC}
        \mathbb E[\|\wu-\wn\|] \leq \frac{2mL}{\lambda n-\widetilde\ell(1+\widetilde\ell/\mu)m} + \sqrt{d_1}\sigma_1,
    \end{equation}
    and
    \begin{equation}\label{eq:vu.CSC}
        \mathbb E[\|\vu-\vn\|] \leq \frac{2mL}{\mu n-\widetilde\ell(1+\widetilde\ell/\lambda)m} + \sqrt{d_2}\sigma_2.
    \end{equation}
    Plugging eq.(\ref{eq:wu.CSC}) and eq.(\ref{eq:vu.CSC}) into eq.(\ref{eq:PD.CSC}) with noise scales given in Lemma \ref{lemma:certi_CSC}, we can get our generalization guarantee:
    \begin{equation}
        \triangle^w(\wu,\vu) \leq \mathcal O \bigg( \big(\frac{L^3 \ell^3 \rho}{\lambda^3 \mu^3} + \frac{L^2 \ell^2}{\lambda^2 \mu} + \frac{L^2 \ell^2}{\lambda \mu^2}\big)\cdot\frac{m^2 \sqrt{d\log(1/\delta)}}{n^2 \epsilon} + \frac{mL^2}{\lambda n} +\frac{mL^2}{\mu n} + \lambda B_{\w}^2 \bigg),
    \end{equation}
    where $d=\max\{d_1,d_2\}$.
\end{proof}

\begin{lemma}[\textbf{Population Strong PD Risk}]
\label{lemma:strrisk_CSC}
    Under the same settings of Lemma \ref{lemma:risk_CSC}, 
    the population strong PD risk for $(\wu,\vu)$ returned by Algorithm \ref{alg:unlearning_CSC} is
    \begin{equation}
        \begin{aligned}
            \triangle^s(\wu,\vu) \leq \mathcal O \bigg( & \big(\frac{L^3 \ell^3 \rho}{\lambda^3 \mu^3} + \frac{L^2 \ell^2}{\lambda^2 \mu} + \frac{L^2 \ell^2}{\lambda \mu^2}\big)\cdot\frac{m^2 \sqrt{d\log(1/\delta)}}{n^2 \epsilon} + \frac{mL^2}{\lambda n} +\frac{mL^2}{\mu n}\\
            &+ \frac{L^2 \ell}{\lambda^{3/2}\mu^{1/2} n} + \frac{L^2 \ell}{\lambda^{1/2} \mu^{3/2} n} + \lambda B_{\w}^2 \bigg),
        \end{aligned}
    \end{equation}
    where $d=\max\{d_1,d_2\}$. In particular, by setting the regularization parameter $\lambda$ as:
    \begin{equation}
    \begin{split}
        \lambda = \max \bigg\{ & \frac{L}{B_{\w}}\sqrt{\frac{m}{n}}, \frac{L\ell m}{B_{\w}\mu n} \big(\frac{\sqrt{d\log(1/\delta)}}{\epsilon}\big)^{1/2}, \big(\frac{L^2\ell^2m^2\sqrt{d\log(1/\delta)}}{B_{\w}^2\mu n^2\epsilon}\big)^{1/3},\\
        & \big( \frac{L^3\ell^3\rho m^2\sqrt{d\log(1/\delta)}}{B_{\w}^2\mu^3n^2\epsilon} \big)^{1/4}, \big( \frac{L^2 \ell}{B_{\w}^2 \mu^{1/2} n} \big)^{2/5}, \frac{1}{\mu} \big( \frac{L^2 \ell}{B_{\w}^2 n} \big)^{2/3} \bigg\}.   
    \end{split}
    \end{equation}
    we have the following population strong PD risk:
    \begin{equation}
    \begin{split}
        \triangle^s(\wu,\vu) \leq \mathcal O \bigg( & c_1 \sqrt{\frac{m}{n}} + c_2\frac{m}{n} + c_3 \big(\frac{\sqrt{d\log(1/\delta)}}{\epsilon}\big)^{1/2}\frac{m}{n} + c_4 \big(\frac{\sqrt{d\log(1/\delta)}}{\epsilon}\big)^{1/3} \big(\frac{m}{n}\big)^{2/3} \\
        & + c_5 \big(\frac{\sqrt{d\log(1/\delta)}}{\epsilon}\big)^{1/4} \sqrt{\frac{m}{n}} + c_6 \frac{1}{n^{2/5}} + c_7 \frac{1}{n^{2/3}} \bigg),
    \end{split}
    \end{equation}
    where $c_1, c_2, c_3, c_4, c_5, c_6$ and $c_7$ are constants that depend only on $L, l, \rho, \mu$ and $B_{\w}$.
\end{lemma}
\begin{proof}
    An application of Lemma \ref{lemma:learning_strPD} gives that
    \begin{equation}\label{eq:CSC_emp_strPD}
        \begin{aligned}
            \mathbb E[\max_{\v\in \mathcal V} \widetilde F(\wn, \v) - \min_{\w\in \mathcal W} \widetilde F(\w, \vn)] \leq & \frac{36\sqrt{2} L^2 (2\ell + \lambda)}{n}\left( \frac{1}{\lambda^{3/2} \mu^{1/2}} + \frac{1}{\lambda^{1/2} \mu^{3/2}} \right) \\
            \leq & \frac{108\sqrt{2} L^2 \ell}{n}\left( \frac{1}{\lambda^{3/2} \mu^{1/2}} + \frac{1}{\lambda^{1/2} \mu^{3/2}} \right).            
        \end{aligned}
    \end{equation}
    Using an application of eq.(\ref{eq:unlearning_strPD}) and eq.(\ref{eq:unlearning_strPD_C}), together with eq.(\ref{eq:CSC_emp_strPD}), eq.(\ref{eq:wu.CSC}), eq.(\ref{eq:vu.CSC}) and the noise scales given in Lemma \ref{lemma:certi_CSC}, we have
    \begin{equation}
        \begin{aligned}
            \triangle^s(\wu,\vu) \leq \mathcal O \bigg( & \big(\frac{L^3 \ell^3 \rho}{\lambda^3 \mu^3} + \frac{L^2 \ell^2}{\lambda^2 \mu} + \frac{L^2 \ell^2}{\lambda \mu^2}\big)\cdot\frac{m^2 \sqrt{d\log(1/\delta)}}{n^2 \epsilon} + \frac{mL^2}{\lambda n} +\frac{mL^2}{\mu n}\\
            &+ \frac{L^2 \ell}{\lambda^{3/2}\mu^{1/2} n} + \frac{L^2 \ell}{\lambda^{1/2} \mu^{3/2} n} + \lambda B_{\w}^2 \bigg).
        \end{aligned}
    \end{equation}
\end{proof}

\begin{lemma}[\textbf{Deletion Capacity}]
    Under the same settings as Lemma \ref{lemma:risk_CSC}, the deletion capacity of Algorithm \ref{alg:unlearning.extension.online} is
    \begin{equation}
        m ^ { A,\bar A} _ {\epsilon,\delta,\gamma}(d_1,d_2,n) \geq c \cdot \frac{n\sqrt{\epsilon}}{(d\log(1/\delta))^{1/4}},
    \end{equation}
    where the constant $c$ depends on $L, l, \rho, \mu$ and $B_{\w}$ and $d = \max\{d_1,d_2\}$.
\end{lemma}
\begin{proof}
    By the definition of deletion capacity, in order to ensure the population PD risk derived in Lemma \ref{lemma:risk_CSC} or Lemma \ref{lemma:strrisk_CSC} is bounded by $\gamma$, it suffices to let $m ^ { A,\bar A} _ {\epsilon,\delta,\gamma}(d_1,d_2,n) \geq c \cdot \frac{n\sqrt{\epsilon}}{(d\log(1/\delta))^{1/4}}$.
\end{proof}

\end{document}